\documentclass{article}

\usepackage{PRIMEarxiv}

\usepackage[utf8]{inputenc} 
\usepackage[T1]{fontenc}    
\usepackage{hyperref}       
\usepackage{url}            
\usepackage{booktabs}       
\usepackage{amsfonts}       
\usepackage{nicefrac}       
\usepackage{microtype}      
\usepackage{lipsum}
\usepackage{fancyhdr}       
\usepackage{graphicx}       
\graphicspath{{media/}}     

\usepackage{natbib}
\usepackage{hyperref}
\usepackage{url}
\usepackage{amsmath}
\usepackage{amssymb}
\usepackage{booktabs}
\usepackage{graphicx}
\usepackage{tikz}
\usepackage{pgfplots}
\pgfplotsset{compat=1.17}
\usetikzlibrary{positioning,arrows.meta}
\usepackage[ruled,vlined]{algorithm2e}
\usepackage{array}
\usepackage{xspace}
\newcommand{\panini}{P\={A}\d{n}ini\xspace}

\usepackage{caption}
\title{PĀṇini: Typed Case-Role Grammars for Feasible-by-Construction Robots 
}

\author{
  Mayukh Debnath \\
  Robotics and Autonomous Systems (AI)\\
  Arizona State University\\
  Tempe, USA\\
  \texttt{mdebnath@asu.edu} \\
   \And
  Manan Roy Choudhury \\
  Department of Artificial Intelligence\\
  Arizona State University\\
  Tempe, USA\\
  \texttt{mroycho1@asu.edu} \\
}

\begin{document}
\maketitle

\begin{abstract}
Symbolic approaches to robot design provide interpretable rules but often overlook continuous geometry, while learned approaches handle geometry but offer limited explanations of their decisions. We introduce a typed production grammar that connects the two through a shared representation of robot skills and morphologies. Its rules use Fillmore-style case roles, parameter inheritance, and specificity-based arbitration, while a geometric realization map translates rule composition into transformations in \(SE(3)\). This allows a single grammar to describe both manipulation skills and robot bodies. We show that the grammar is substantially more compact than an untyped alternative and that its rule arbitration can be recovered from learned geometric containment, achieving \(0.905\pm0.062\) decision agreement across 10 seeds, compared with \(0.611\pm0.090\) for a Euclidean control. We also characterize how inheritance affects the compositional properties of the geometric realization map by measuring its homomorphism defect. For morphology optimization, we introduce \panini (\emph{P}roduction-rule \emph{A}nd role-typed compositio\emph{N} for \emph{I}ntelligent, \emph{N}ovel robot des\emph{I}gn), a cross-entropy search method operating directly on the grammar's typed genome. Across 30 methods and 24 seeds, \panini achieves the best mean rank, but is statistically indistinguishable from strong generic optimizers, including IPOP-CMA-ES, CEM, PSO, and TuRBO. Evaluating final designs with reinforcement-learning-trained controllers leads to a similar conclusion: the grammar-based search does not consistently outperform generic optimization. Further search mechanisms designed to exploit grammar structure also fail to establish a general advantage. A curriculum over grammar depth is a promising exception, becoming significantly better than six of seven strong baselines at 48 seeds, although this result requires caution because multiple variants were explored. The grammar's main advantage is therefore not superior optimization performance, but structured and feasible design generation. Every generated candidate satisfies the grammar's structural constraints, whereas unstructured methods achieve legality rates of only 0--82\%. Coordinated limb placement maintains 95\% joint feasibility in a highly constrained design space, compared with 55\% for range-constrained resampling and no feasible designs from 6,000 flat-parameter samples. The grammar also supports designs with different limb counts, which the compared alternatives do not reach. We further examine the limits of these guarantees: structural legality does not ensure locomotion, and geometric feasibility does not reliably predict simulator failure. Together, these results establish typed production grammars as a compact, interpretable, and constraint-aware framework for robot skill and morphology generation, while showing that their structural advantages do not automatically translate into better black-box optimization. Codes for every experiment in this paper are available at \texttt{https://github.com/mdebnath22/panini}.
\end{abstract}

\keywords{Robot Design \and Typed Production Grammars \and Morphology Generation \and Robot Skill Synthesis \and Geometric Reasoning \and Case-Role Representation \and Grammar-Based Optimization \and Feasibility-Constrained Search \and Black-Box Optimization \and Reinforcement Learning}

\section{Introduction}
\label{sec:introduction}

Black-box optimizers are routinely applied to structured design spaces such as robot bodies, where each candidate must be both \emph{good} and \emph{legal}. Unconstrained continuous or graph-based parameterizations can express a wide range of bodies, but whether a body is valid must be checked after it is generated rather than guaranteed by construction. This also affects recent LLM-based low-level design editors \citep{Qiu2024,Song2025,Qiu2025}. The alternative is to represent a body as a structured object governed by a discrete, checkable grammar \citep{Sims1994}, echoing the older trade-off in planning between explainable symbolic operators \citep{Fikes1971} and reactive loops that gain robustness by dropping explicit structure \citep{Brooks1991}. This raises a question that is rarely separated empirically: when a black-box optimizer searches a structured space, does the structure help it find \emph{better} designs, or only restrict which designs it can propose?

Existing grammar-based approaches address parts of this. RoboGrammar \citep{Zhao2020} and GLSO \citep{Hu2022} use graph grammars to constrain robot morphology, but their rules lack checkable geometric semantics and are typed only by graph adjacency. Region and order embeddings show that asymmetric relations such as entailment or default-versus-override can be represented geometrically as containment \citep{Vilnis2015,Vendrov2015,RenQuery2Box2020,Ganea2018}, but to our knowledge this has not been tested as a way to resolve competing rules inside a grammar.

We build a typed production grammar that combines semantic roles, parameter inheritance, and specificity-based rule arbitration. For manipulation, rules carry Fillmore-style case roles \citep{Fillmore1968} (Agent, Patient, Instrument, Location, Source, Goal); for robot morphology, a separate structural role set (Root, Limb, Segment, Joint, Terminal). A geometric realization map turns every valid derivation into a sequence of rigid-body transformations in $SE(3)$. Rather than assuming that this map preserves the algebraic structure of rule composition, we measure its composition error over sampled rule pairs and longer chains. We then evaluate the grammar as a search space for morphology optimization, asking two separate questions: does it make search \emph{find better bodies}, and does it make the bodies it proposes \emph{structurally valid}? The answers differ, and we report both as measured.

\paragraph{What we find.}
\begin{itemize}
  \item \textbf{Structure guarantees feasibility.} Every candidate the grammar produces is legal by construction, against $0$--$82\%$ for unstructured edits. In a very-low-feasibility space, coordinated limb placement keeps $95\%$ of edits jointly feasible, range-constrained resampling keeps $55\%$, and flat parameterization finds no feasible design in $6{,}000$ draws. Only the grammar reaches designs that require a different limb count.
  \item \textbf{Structure does not improve fitness over strong generic optimizers.} A grammar-guided cross-entropy search has the best mean rank among 30 methods at $n=24$ seeds but is statistically indistinguishable from IPOP-CMA-ES, CEM, PSO, and TuRBO (Nemenyi $p=1.00$). Re-scoring each method's final designs with RL-trained controllers does not favor the grammar either (Friedman $p=0.71$ in a clean six-method re-run).
  \item \textbf{Mechanisms that give the grammar a role in the search dynamics mostly do not help.} None separates from the strongest generic optimizers, and transfer to held-out limb counts is significantly worse. The exception is a curriculum over grammar depth, which is marginally better at $n=24$ and better than six of seven strong baselines at $n=48$, although not every comparison survives correction for the number of variants tried.
  \item \textbf{The grammar's internal semantics are measurable.} The realization map is a monoid homomorphism up to a defect that we measure when inheritance is active, rule arbitration is recoverable as learned geometric containment ($0.905\pm0.062$ decision agreement over 10 seeds, versus $0.611\pm0.090$ for a Euclidean control), and the grammar is far more compact than an untyped alternative.
\end{itemize}

We read the consistent null on fitness as evidence that well-tuned generic optimizers are hard to beat on this landscape however the grammar enters the search. Both guarantees have limits that we also measure: a legal body is not thereby a locomoting one, and the geometric feasibility gate is a design-specification filter that does not predict simulator failure.

\paragraph{Contributions.}
\begin{itemize}
  \item A typed production grammar that unifies semantic and structural case roles, parameter inheritance, specificity-based arbitration, and a geometric realization map into $SE(3)$, covering manipulation skills and robot morphology under one formalism.
  \item \textbf{Structure does not improve fitness over strong generic optimizers.} \panini, a grammar-guided cross-entropy search, has the best mean rank among 30 methods at $n=24$ seeds but is statistically indistinguishable from IPOP-CMA-ES, CEM, PSO, and TuRBO (Nemenyi $p=1.00$).
  \item \panini, benchmarked against 29 alternatives in a controlled comparison that separates what structure buys (feasibility by construction) from what it does not (fitness, or RL-scored quality, over strong generic optimizers).
\end{itemize}

\section{Method}
\label{sec:method}

\subsection{A Concrete Example and Notation}
\label{sec:example}

A robot with gripper $g$ must move block $b$ into bin $k$ on table $t$ by approaching, grasping, lifting, transporting, and placing it. An approach height declared in the first step must be available to every later step that needs it, without being restated (\emph{inheritance}). If the block has an attached tool, a more specific approach rule should take precedence over the generic rule without removing it (\emph{specificity-based arbitration}). Table~\ref{tab:notation} defines the notation used throughout.

\begin{table}[h]
\caption{Notation used throughout the paper.}
\label{tab:notation}
\centering
\small
\begin{tabular}{@{}>{\raggedright\arraybackslash}p{0.13\linewidth} >{\raggedright\arraybackslash}p{0.80\linewidth}@{}}
\toprule
Symbol & Meaning \\
\midrule
$\mathcal{R}$ & case-role set $\{$Agent, Patient, Instrument, Location, Source, Goal$\}$ \\
$F$ & case frame: partial map $\mathcal{R} \to$ Entities \\
$r$ & production rule (condition, transformation, parameters) \\
$s(r)$ & specificity index of rule $r$ \\
$T$ & derivation: ordered sequence of applied rules \\
$\Phi(T)$ & $SE(3)$ transform (or kinematic tree) realizing derivation $T$ \\
$d_{geo}$ & left-invariant geodesic distance on $SE(3)$ \\
$n_\ell$, $k$ & number of legs (limbs) and segments per leg of a robot body \\
$n$ & sample size of a statistical comparison (random seeds or trials), never a leg count \\
$n_{\text{inst}}$ & number of randomly sampled concrete instantiations tested per abstract rule-pair or chain type (role bindings, parameters, scenes), never a seed count \\
\bottomrule
\end{tabular}
\end{table}

\subsection{Case-Role Schema}
\label{sec:schema}

Six case roles, following \cite{Fillmore1968}: Agent, Patient, Instrument, Location, Source, Goal. A case frame assigns roles to scene entities, each carrying a pose in $SE(3)$. Robot body design (Section~\ref{sec:morphology}) reuses this mechanism with a separate set of roles for body parts: Root, Limb, Segment, Joint, and Terminal. The role sets are disjoint because they apply to categorically different kinds of entities.

\subsection{Production Rules, Inheritance, and Arbitration}
\label{sec:rules}

A production rule $r$ is $(\text{cond}, \text{req}, \text{params}, \text{emit}, \text{consume}, \text{block}, s)$: a boolean condition, required case roles, default parameters, STRIPS-style markers for legality gating, and a specificity index $s(r) \in \mathbb{N}$.

\textbf{Inheritance.} A parameter in $r$'s inheritance-export set is written into the shared derivation state. It stays available to every later rule, unrestated, until overridden. Inheritance is the only channel for cross-rule information flow. That is also why it is the only possible source of departure from strict compositionality in the geometric realization map $\Phi$ (Section~\ref{sec:results-homomorphism}).

\textbf{Specificity arbitration.} When several rules match, the most specific one fires; if more than one rule ties for most specific, the one declared first wins. A losing rule is not deleted. It remains available wherever no more specific rule applies. This default-versus-override relation is what Section~\ref{sec:boxes} tests as geometric containment.

Algorithm~\ref{alg:derive} (Appendix~\ref{app:pitfalls}) specifies the derivation procedure exactly. Given a plan -- an ordered sequence of intended operations with role bindings -- it resolves, for each step, which rule fires and with which parameters: the most specific match wins, ties are broken by declaration order, and an inherited value overrides the matched rule's own default unless the step supplies an explicit override, which wins over both. The result is the full derivation that $\Phi$ subsequently realizes geometrically.

\subsection{The Geometric Realization Map $\Phi$}
\label{sec:phi}

$\Phi$ sends each applied rule to a screw motion -- a twist $\xi \in \mathfrak{se}(3)$ executed as $\exp(s\hat\xi)$, $s \in [0,1]$ -- and a derivation to the composition of its rules' twists:
\begin{equation}
\Phi(r_1 \circ \cdots \circ r_n) = \exp(\hat\xi_1)\cdots\exp(\hat\xi_n).
\label{eq:phi}
\end{equation}
For manipulation, each rule's twist follows in closed form from its role bindings. For robot bodies, the same mechanism places each part's frame relative to its parent, yielding a kinematic tree directly (Section~\ref{sec:morphology}). We claim $\Phi$ is a monoid homomorphism, and Section~\ref{sec:results-homomorphism} tests this claim rather than assuming it, since inheritance can make a rule's realized twist depend on more than that rule alone.

To state this precisely: let $\mathcal{D}$ be the free monoid of applied-rule sequences under concatenation ($\circ$), and let $\Phi: \mathcal{D} \to SE(3)$ be given by Equation~\ref{eq:phi}. $\Phi$ is a monoid homomorphism on a pair $(r_1, r_2)$ -- $\Phi(r_1 \circ r_2) = \Phi(r_1)\cdot\Phi(r_2)$ exactly, in $SE(3)$'s own group operation -- exactly when $r_2$'s resolved parameters $\theta_2$ (the inheritance step in Algorithm~\ref{alg:derive}) do not depend on any value $r_1$ exported through inheritance; we call such a pair \emph{inheritance-inactive}. When $r_2$ instead imports a parameter from $r_1$'s export set, its twist $\xi_2$ becomes a function of $r_1$'s resolved parameters, and the composition identity must be checked rather than assumed, not assumed to hold. We do not prove a bound on this gap holding for every derivation the grammar can produce; Section~\ref{sec:results-homomorphism} instead reports its measured value on every inheritance-active and inheritance-inactive pair and chain sampled, an empirical characterization rather than a general theorem.

\textbf{Measuring the gap.} Every element $T \in SE(3)$ decomposes uniquely as $T = (R, p)$, a rotation $R \in SO(3)$ and a translation $p \in \mathbb{R}^3$. Composition quality between any two transforms $T_1=(R_1,p_1)$ and $T_2=(R_2,p_2)$ is measured with the standard left-invariant geodesic distance on $SE(3)$, which separates rotational from translational discrepancy:
\begin{equation}
d_{geo}(T_1, T_2) = \sqrt{\lVert \log_{SO(3)}(R_1^\top R_2) \rVert^2 + w \, \lVert p_2 - p_1 \rVert^2},
\label{eq:dgeo}
\end{equation}
with $w{=}1$ throughout. Since this mixes radians and meters into one scalar, Section~\ref{sec:results-homomorphism} also reports the two components separately.

Homomorphism consistency is then this same distance applied to one specific pair of transforms: the realization of a composed derivation, against the composition of its parts' separate realizations,
\begin{equation}
\delta(r_1, r_2) = d_{geo}\big(\underbrace{\Phi(r_1{\circ}r_2)}_{T_1},\ \underbrace{\Phi(r_1){\cdot}\Phi(r_2)}_{T_2}\big),
\label{eq:defect}
\end{equation}
with $r_2$'s inheritance severed on the right-hand side (so $T_2$ is computed as if $r_2$ never saw $r_1$'s exports), and extended to full chains of length $k=2,\ldots,6$ by severing inheritance at every step. $\delta=0$ recovers exact homomorphism for that pair or chain; $\delta>0$ is the defect Section~\ref{sec:results-homomorphism} measures.

\subsection{Extension to Robot Morphology}
\label{sec:morphology}

An identical grammar generates robot bodies, using the disjoint role set of Section~\ref{sec:schema} and a seven-rule catalogue: AddRoot, AddLimb (with a specificity-2 override variant), Articulate, Extend (inheriting the current segment template), Terminate, and Symmetrize. Each AddLimb application mounts one limb on the torso. Applied $n_\ell$ times in a row (once per leg, where $n_\ell$ is the genome's leg count, \S\ref{sec:panini-method}), with the $i$-th application placing its mount angle at $2\pi i/n_\ell$ around the torso's circumference (a deterministic function of the application index and $n_\ell$, up to the genome's bounded mount-angle jitter), it produces an $n_\ell$-legged body with uniform angular spacing by construction, not by a post-hoc check that rejects unevenly spaced bodies. Segment geometry is declared once and inherited by every later Extend within a limb, so an $n_\ell$-legged, $k$-segment body uses the same rules regardless of $n_\ell$ and $k$ (Section~\ref{sec:compactness}). The full catalogue is in Appendix~\ref{app:rules}.

\subsection{Geometric Feasibility Layer}
\label{sec:feasibility}

Before a derived body is compiled and simulated, five geometric constraints are checked against a task specification: workspace reach, ground clearance, mass budget, support-polygon area, and minimum inter-terminal separation. A body failing any constraint is rejected before simulation. Section~\ref{sec:results-gate} tests what this buys and what it does not.

\subsection{Searching the Grammar's Genome}
\label{sec:panini-method}

A morphology genome has five fields: leg count $n_\ell\in[2,8]$, segments per leg $k\in[1,3]$, segment length and radius, and mount-angle jitter. Every genome is derived through the grammar, realized by $\Phi$, and gated (\S\ref{sec:feasibility}) before simulation. Two grammar-based searchers operate on it. Grammar-guided evolution (GGE) is a $(\mu+\lambda)$ loop that mutates one parent at a time. \panini{} replaces this with cross-entropy resampling \citep{Rubinstein1999}. Each generation fits a diagonal Gaussian to the top 30\% of a population of 10, carries the two best genomes forward unchanged, and rounds the discrete fields when the genome is built; the last 10\% of the budget is a trust-region refinement around the incumbent. The standard deviation of $k$ is floored at $0.6$. Without this floor the elite refit collapsed onto $k=2$ in all 24 seeds; with it, the number of distinct $(n_\ell,k)$ combinations reached rises from 3 to 6, and mean fitness changes from $0.403$ to $0.411$. This default configuration is not an arbitrary setting: a sweep over population size, elite fraction, elitism, and refinement fraction left every variant within $0.011$ mean fitness of the default (paired $p=0.57$--$0.96$, $n=24$, budget 60).

Because every baseline in \S\ref{sec:setup-search} also searches this genome through the same derivation pipeline, the grammar defines the search space for \emph{all} methods. \panini differs from them only in optimizer mechanics. Section~\ref{sec:edge} tests variants in which the grammar also takes part in the search dynamics.

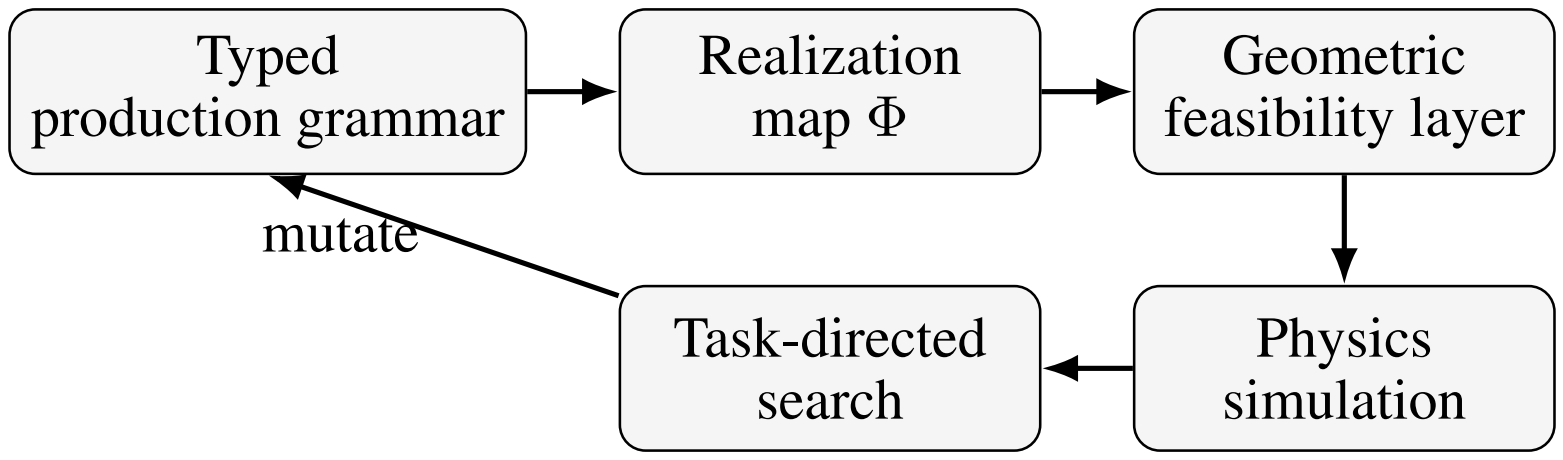
\begin{figure}[h]
\centering
\resizebox{0.62\linewidth}{!}{\begin{tikzpicture}[
  box/.style={draw, rounded corners, minimum width=2.3cm, minimum height=0.9cm, align=center, font=\small, fill=gray!8},
  arr/.style={-{Latex[length=2mm]}, thick}
]
\node[box] (grammar) {Typed\\production grammar};
\node[box, right=0.5cm of grammar] (phi) {Realization\\map $\Phi$};
\node[box, right=0.5cm of phi] (gate) {Geometric\\feasibility layer};
\node[box, below=0.6cm of gate] (phys) {Physics\\simulation};
\node[box, below=0.6cm of phi] (search) {Task-directed\\search};
\draw[arr] (grammar) -- (phi);
\draw[arr] (phi) -- (gate);
\draw[arr] (gate) -- (phys);
\draw[arr] (phys) -- (search);
\draw[arr] (search) -- (grammar.south) node[midway, left, font=\small] {mutate};
\end{tikzpicture}}
\caption{A derivation is realized by $\Phi$, gated for feasibility before any physics call, evaluated in simulation, and the resulting fitness drives mutation of the grammar's own parameters.}
\label{fig:pipeline}
\end{figure}

A schematic of the full pipeline -- grammar, realization map $\Phi$, feasibility gate, physics, and search closing the loop back onto the grammar's parameters -- is given as Figure~\ref{fig:pipeline}.

\section{Experimental Setup}
\label{sec:setup}

\subsection{Rule Catalogues, Task Specifications, and the Simulated Body}
\label{sec:setup-catalogues}

\begin{figure}[h]
\centering
\begin{minipage}{0.97\linewidth}
\centering
\includegraphics[width=0.95\linewidth]{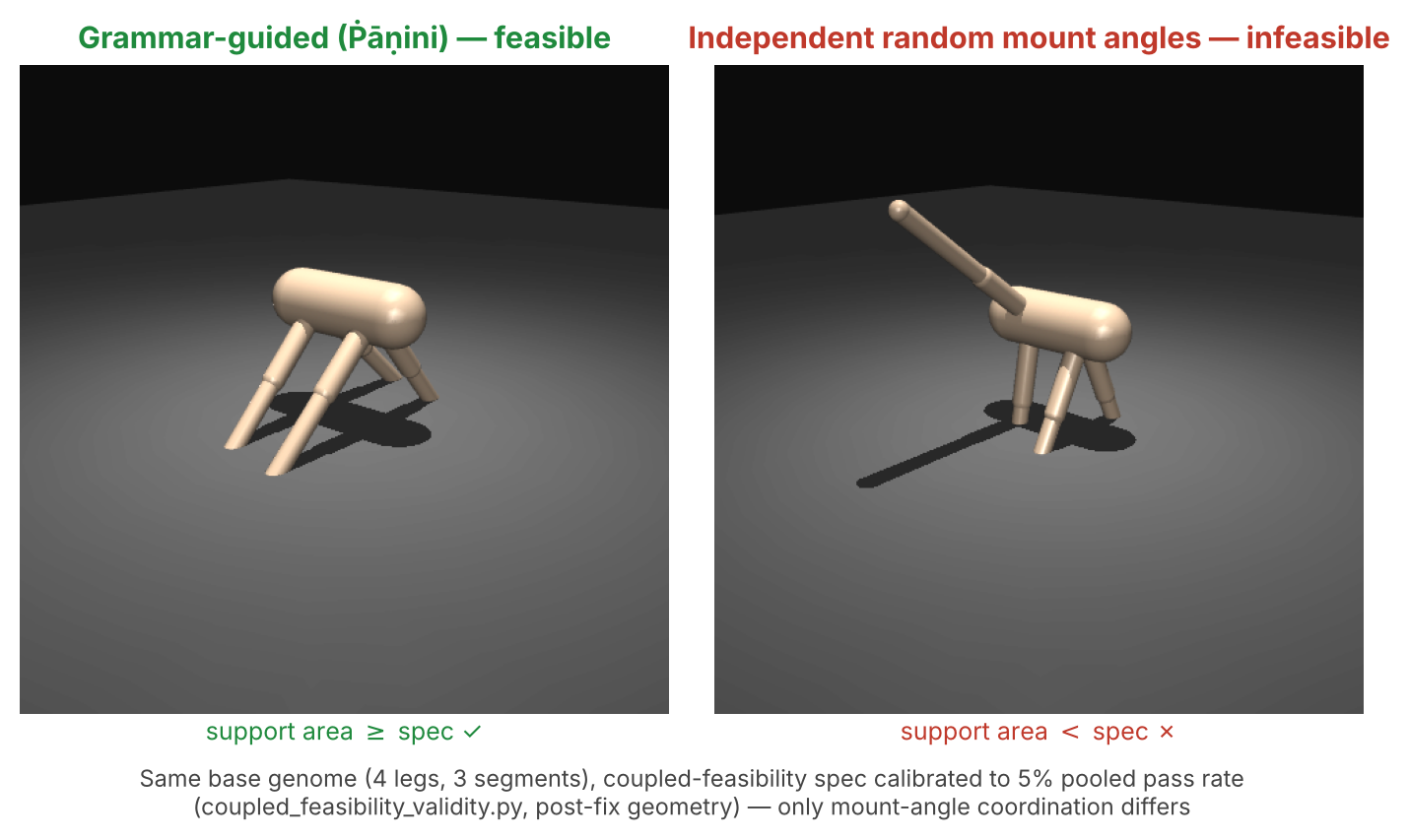}
\\[2pt]
\footnotesize (a) Same base genome (4 legs, 3 segments/leg); only mount-angle coordination differs. The \panini design is feasible under a coupled-feasibility spec calibrated to a 5\% pooled pass rate; the independent-random-angle counterfactual is not.
\end{minipage}
\hfill
\begin{minipage}{0.67\linewidth}
\centering
\includegraphics[width=0.6\linewidth]{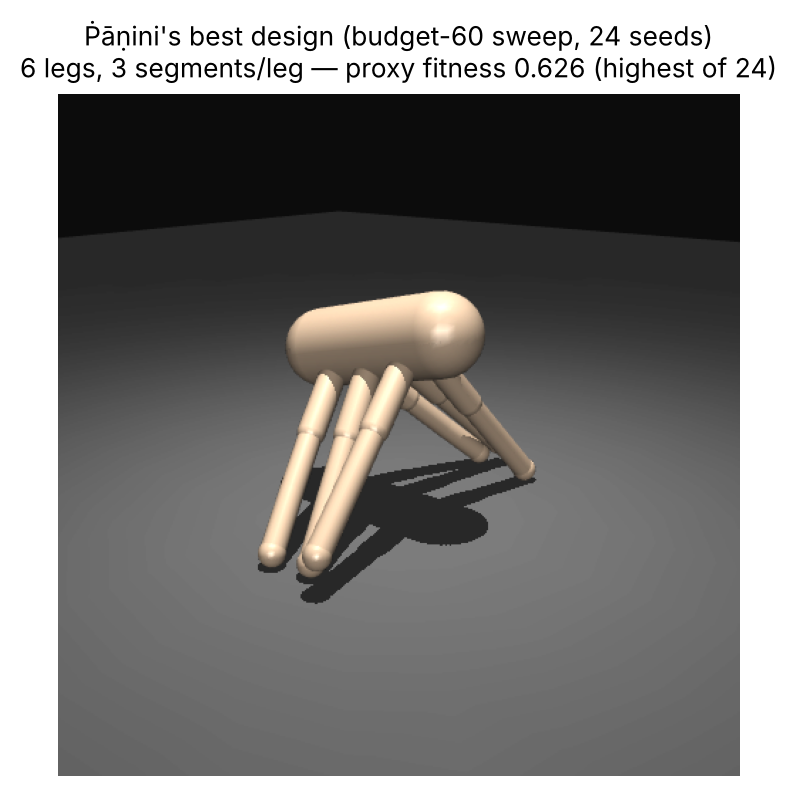}
\\[2pt]
\footnotesize (b) \panini's best design from the budget-60 sweep (24 seeds): 6 legs, 3 segments/leg, proxy fitness 0.626 (highest of the 24 seeds).
\end{minipage}
\caption{Feasibility by construction (a) and the best found design (b).}
\label{fig:panini}
\end{figure}

The manipulation grammar (8 rules: Approach in two specificity variants, Grasp, Lift, Transport, Place, RotateTool, Retract) and the morphology grammar (7 rules, \S\ref{sec:morphology}) were built independently. Both are given in full in Appendix~\ref{app:rules}. The morphology task spec used throughout is forward locomotion. Fitness is center-of-mass displacement under a fixed sinusoidal control signal over a fixed horizon: a controller-agnostic proxy \citep{Sims1994}, chosen so that morphology search is not confounded with control-policy search. Every compiled body's torso is attached to the world by a six-degree-of-freedom free joint, genuinely free to translate and rotate under gravity, with per-joint damping, armature, and an implicit integrator for numerical stability under contact; on 100 freshly sampled 8-limb bodies this produces 0 numerical blowups and 0 degenerate (zero-actuator) bodies, with displacement over a 2-second horizon ranging $-0.34$ to $+0.88$\,m (mean $0.13$\,m), a physically plausible range for a body of this scale. The feasibility specification (\S\ref{sec:feasibility}) is calibrated so that $\approx7\%$ of fresh random genomes pass the geometric check ($6.8\%$ of 500, for the gate ablation of \S\ref{sec:results-gate}) -- a design-specification pass rate, not a simulator-failure rate (\S\ref{sec:results-gate}). Whether the proxy predicts RL-trained performance is tested in Appendix~\ref{app:proxy-rl}: on a catalogue rebuilt after a limb-mounting geometry fix, the rank correlation is positive but borderline (Spearman $\rho=0.355$, $p=0.050$, $n=31$), so the proxy is at best a weak predictor. Figure~\ref{fig:panini} illustrates the construction contrast directly: panel (a) shows an independently-sampled mount-angle configuration that fails our coupled-feasibility spec while the grammar-guided design from the same base genome passes it; panel (b) shows \panini's best design from the 24-seed budget sweep.

\subsection{Search Baselines}
\label{sec:setup-search}

\panini and GGE (\S\ref{sec:panini-method}) are compared against 28 alternatives under one accounting rule: one fitness query is one unit of budget, whatever the method's population size or update schedule. Twenty are classical, self-adaptive, or quality-diversity black-box optimizers: random search, PSO \citep{Kennedy1995}, simplified and IPOP \citep{Hansen2016} CMA-ES, separable NES \citep{Wierstra2014}, hill-climbing, CEM \citep{Rubinstein1999}, Bayesian optimization \citep{Snoek2012}, DE/rand/1/bin \citep{Storn1997}, SHADE \citep{Tanabe2013}, augmented random search \citep{Mania2018}, MAP-Elites \citep{MouretClune2015}, novelty search \citep{Lehman2011}, simulated annealing, Nelder-Mead, $(1{+}1)$-ES, SPSA \citep{Spall1992}, coordinate descent, LHS with greedy refinement \citep{McKay1979}, and TuRBO \citep{Eriksson2019}. Seven more are budget-matched reimplementations of recent morphology-design search mechanisms (\S\ref{sec:recent-sota}), plus one intermediate grammar-based variant (GGE-v2). Section~\ref{sec:edge}'s targeted experiments use a fixed \emph{reference cluster} of seven non-grammar methods: the five ranking directly below \panini at budget 60 (IPOP-CMA-ES, CEM, PSO, coarse-to-fine, TuRBO), LHS+refinement, and the simplified CMA-ES, the cluster's weak tail.

\subsection{Validity Baselines}
\label{sec:setup-validity}

Grammar-guided mutation (edit one genome parameter, re-derive) is compared against 14 alternatives spanning three families. \emph{Direct representation edits} at varying severity: an adversarial mixture (15\% collapse-to-zero, 15\% sign flip, 15\% five-to-fifty-fold blow-up, 55\% moderate rescale), mild-only and extreme-only variants of it, structure-blind uniform resampling, deletion/duplication of a compiled geometry element, naive subtree-swap crossover, a raw byte-level bit-flip, and the mixture compounded at 5 and 10 mutations instead of 3. \emph{Structure-blind but range-aware resampling}: per-attribute-type plausible-range resampling (e.g.\ radii constrained positive) with no cross-field awareness, naive numeric-average crossover, and pure re-initialization through the grammar (a sanity floor, since it reflects the grammar's own guarantee). \emph{A trained-but-unconstrained latent perturbation}: a minimal linear autoencoder (PCA) fit to grammar-generated valid designs, perturbed in latent space and decoded -- a simplified stand-in for the GLSO-style contrast between learned-space and construction-guaranteed validity. Validity requires the compiled model to load in MuJoCo \citep{Todorov2012} and step forward without non-finite state.

\subsection{Metrics and Statistical Protocol}
\label{sec:setup-metrics}

Homomorphism consistency is measured with the defect $\delta$ of Equation~\ref{eq:defect} (\S\ref{sec:phi}), on sampled rule pairs and on chains of length $k=2,\ldots,6$. Compactness is measured as rule count under three conditions -- fully flat, typed without inheritance, and the full grammar -- across a grid of 2--10 legs and 1--4 segments per leg. Specificity-as-containment is measured as agreement between a learned region embedding's smallest-containing-box decision and the symbolic arbitration decision, on held-out states. For every multi-method search comparison we report per-method mean and standard deviation across seeds, then a Friedman test followed, where the omnibus test is significant, by a Nemenyi post-hoc test \citep{Demsar2006}; with 30 simultaneous methods, uncorrected pairwise tests would produce spurious significance. We treat $n=24$ seeds as the decisive sample size throughout. The one result that needed more power was rerun at $n=48$, and both sample sizes are reported. Paired $t$-tests with 95\% confidence intervals appear only as uncorrected, secondary evidence and are labelled as such. For the validity comparison, we report raw pass rate out of 100 trials per condition. All results use one pinned simulator version; Appendix~\ref{sec:repro} reports how sensitive the search statistics are to that choice.

\section{Results}
\label{sec:results}

We test whether the grammar composes geometry consistently, reduces rule counts, generalizes to unseen compositions, supports learned rule selection, improves morphology search, and preserves structural validity. We distinguish these structural benefits from improvements in task performance.

\textbf{Compositional consistency of $\Phi$.}
\label{sec:results-homomorphism}
Does realizing a derivation match composing its separately realized parts (Eq.~\ref{eq:defect})? We tested both independently built catalogues with $n_{\text{inst}}=50$--$60$ instances per pair and chains of length two through six. These are instance counts, not seed counts (Table~\ref{tab:notation}); \emph{the results describe sampled derivations, not a general proof}.

\emph{Composition is exact without active inheritance, with small defects when inheritance is active.} The defect $\delta$ was zero for every sampled inheritance-inactive pair and $0.037$--$0.038$ for adjacent inheritance-active pairs. Across full chains, it ranged from $0.000$ to $0.081$. Rather than growing steadily with length, it tracked whether the chain ended with Extend, which reads inherited segment length, or Articulate, which does not. For Articulate-then-Extend, the defect was entirely translational: $0.0^\circ$ rotation and $3.75\pm2.08$\,cm translation, consistent with Extend adding a segment by translation. Translating the Patient's pose also translated the realized motion by the same group element, with zero deviation in every sampled case.

\emph{Exact composition holds when the second rule does not use the first rule's inheritance exports.} Whenever $r_2$ uses no parameter exported through inheritance by $r_1$,
$\Phi(r_1{\circ}r_2)=\Phi(r_1){\cdot}\Phi(r_2)$.
Under this condition, $r_2$ receives identical parameters whether evaluated jointly or independently.

\textbf{Grammar compactness.}
\label{sec:compactness}
\emph{Typing reduces rule counts, and inheritance provides a further reduction.} Typing reduces counts by $2.1$--$4.1\times$, and inheritance adds another $1.3$--$2.0\times$ reduction (Figure~\ref{fig:compactness}, Appendix~\ref{app:figs}). For 4-leg$\times$2-seg, 6-leg$\times$3-seg, and 8-leg$\times$3-seg bodies, flat counts of 17/37/49 become 8/10/12 with typing and 6/6/6 with inheritance. The manipulation grammar represents 30 variants with 5 rules rather than 80 ($16\times$).

\emph{The compactness advantage grows with body size.} Flat counts grow linearly, while the full typed grammar remains at six rules, reaching a $13.5\times$ advantage for the largest tested body. At the smallest, with 2 legs and 1 segment, the ratio is $0.83\times$: fixed grammar overhead slightly exceeds the flat representation. The combinatorial bound $(\text{max legs}-1)\times\text{max segments}$ gives 21 for legs in $[2,8]$ and segments in $[1,3]$, matching enumeration.

\emph{Inheritance executes across all limbs, not just the first.} A structure-aware trace shows that all limbs in $100/100$ mutated bodies used the genome's radius and length when these overrode rule defaults (Appendix~\ref{sec:inheritance-ablation}). Without inheritance, no complete body did ($0/100$), and only 8 of 415 non-first limbs matched by chance. This test is separate from structural validity: MuJoCo can accept a body even when its parameters differ from those intended.

\textbf{Compositional generalization.}
\label{sec:compgen}
We trained a transformer $\Phi_\theta$ on 600 derivations per domain using a joint-embedding predictive objective \citep{LeCun2022,Assran2023}. Held-out tests use unseen primitive orderings for manipulation and 7--8 legs with 3 segments for morphology, beyond training's 3--6 legs and 1--2 segments.

\emph{Rule identity generalizes perfectly, but geometry prediction does not.} Rule identity is recovered perfectly on both splits in both domains (Table~\ref{tab:compgen}), while morphology geometry error increases substantially on held-out bodies. Manipulation error instead falls, but this reflects template composition, not better generalization: training includes a difficult rotation operation with a randomly sampled axis that the held-out split lacks. The diagnostic rules out template length as the explanation (Appendix~\ref{sec:compgen-diagnostic}).

\begin{table}[h]
\caption{Compositional generalization in both domains.}
\label{tab:compgen}
\centering
\begin{tabular}{llcc}
\toprule
Domain & Split & Error & Rule-ID accuracy \\
\midrule
Manip. & In-dist. & 0.044 (twist MSE) & 100\% \\
Manip. & Held-out & 0.012 (twist MSE) & 100\% \\
Morph. & In-dist. & 0.0002 (placement MSE) & 100\% \\
Morph. & Held-out & 0.043 (placement MSE) & 100\% \\
\bottomrule
\end{tabular}
\end{table}

\textbf{A physically simulated manipulation benchmark.}
\label{sec:manip-benchmark}
The preceding manipulation tests are symbolic. We therefore tested physical execution in a minimal pick-and-place benchmark against a location-blind planner and an object-aware planner given the same ground-truth positions but without staged hover/grasp/lift ordering (Appendix~\ref{app:manip-benchmark}).

\emph{The grammar succeeds physically, but does not outperform object-aware planning.} It beats the blind planner ($100\%$ versus $0\%$ grasp success) but ties the object-aware planner ($100\%$ each on every task), which is slightly better on raw reach distance. This establishes successful physical execution, not an advantage for staging over competent object-aware planning. It remains a proof of concept rather than an evaluation as extensive as the morphology tests below.

\textbf{Specificity as learned geometric containment.}
\label{sec:boxes}
Can learned geometry reproduce symbolic specificity-based arbitration (Section~\ref{sec:rules})? We evaluated containment and winner agreement across 10 training seeds and $300$--$400$ held-out states (Appendix~\ref{app:figs}).

\emph{Learned containment is exact, including on held-out transitive pairs.} The untrained control has agreement $0.000$. Trained containment is $1.000$ in the single-pair-per-grammar setting, across four width-floor settings, and on held-out transitive pairs, at every seed.

\emph{Winner selection is less exact, but containment and ordering outperform proximity.} Box containment achieves $0.905\pm0.062$ agreement (range $0.812$--$1.000$), compared with $0.938\pm0.038$ for order embeddings \citep{Vendrov2015} and $0.611\pm0.090$ for nearest prototypes (range $0.448$--$0.719$). Containment also holds across four regularization settings on four-level chains in two rule families designed to separate specificity from feature overlap.

\emph{Noise changes rule selection without breaking legality.} Perturbing specificity indices causes no legality violations in 800 derivations: required-role and marker gating enforce legality independently. Winner changes rise through $0$, $2.5$, $8.0$, and $12.0\%$ across noise levels. Independent feature-bit flips yield agreement of $0.812$, $0.812$, $0.719$, $0.562$, $0.438$, and $0.344$ as flip probability rises from $0$ to $0.3$, showing gradual degradation rather than sudden collapse (Figure~\ref{fig:noise}).

\textbf{Task-directed morphology search.}
\label{sec:results-search}
\emph{\panini has the best numerical results, but no resolved lead over strong generic optimizers.} Across 30 methods, 24 seeds, and 60 evaluations, it has the best mean rank ($9.33$) and fitness ($0.411\pm0.088$; Table~\ref{tab:search30}). The overall test detects differences ($\chi^2=134.15$, $p=1.8\times10^{-15}$), yet Nemenyi separates \panini only from the five weakest methods.

Uncorrected paired comparisons agree. The advantage over IPOP-CMA-ES is $+0.013$ ($p=0.64$); advantages over CEM, PSO, coarse-to-fine, and TuRBO are $+0.021$--$+0.050$, with every confidence interval including zero ($p=0.20$--$0.38$). \emph{The comparisons clear the earlier non-inferiority margin but do not establish equivalence}: lower bounds exceed $-0.05$, while upper bounds do not fall below $+0.05$.

\emph{The numerical lead also carries a selection caveat.} \panini borrows mechanisms from its competitors, and its configuration was selected after earlier task comparisons (Section~\ref{sec:panini-method}). GGE ranks 9th ($0.349\pm0.130$), separates only from the two weakest methods, and trails \panini by $0.062$ ($p=0.065$). Full results use the corrected body geometry and appear in Appendix~\ref{sec:search-full}.

\begin{table}[h]
\caption{Search comparison: 30 methods, budget 60, and 24 seeds, sorted by mean rank. Nemenyi $p$-values compare each method with \panini. Full results: Table~\ref{tab:search30-full}.}
\label{tab:search30}
\centering
\small
\begin{tabular}{lccc}
\toprule
Method & Mean $\pm$ std. & Mean rank (of 30) & Nemenyi $p$ \\
\midrule
\panini & $0.411\pm0.088$ & 9.33 & --- \\
IPOP-CMA-ES \citep{Hansen2016} & $0.397\pm0.086$ & 10.04 & 1.000 \\
Cross-entropy method \citep{Rubinstein1999} & $0.390\pm0.104$ & 10.88 & 1.000 \\
PSO \citep{Kennedy1995} & $0.374\pm0.138$ & 11.19 & 1.000 \\
Coarse-to-fine (RoboMoRe analogue) & $0.383\pm0.095$ & 12.04 & 1.000 \\
TuRBO \citep{Eriksson2019} & $0.361\pm0.144$ & 12.17 & 1.000 \\
\multicolumn{4}{c}{\emph{(GGE-v2 and augmented random search: ranks 12.58--13.25, $p=1.000$)}} \\
Grammar-guided evolution (GGE) & $0.349\pm0.130$ & 13.33 & 1.000 \\
\multicolumn{4}{c}{\emph{(16 further methods: ranks 13.56--18.48, all $p\ge0.08$)}} \\
\multicolumn{4}{c}{\emph{(Simplified CMA-ES, coordinate descent, fine-tuned controller:}} \\
\multicolumn{4}{c}{\emph{ranks 18.98--20.58, $p\le0.043$)}} \\
Archive criterion-rotation (Debate2Create) & $0.202\pm0.123$ & 23.58 & $<10^{-4}$ \\
Nelder-Mead & $-0.054\pm0.315$ & 26.46 & $<10^{-4}$ \\
\bottomrule
\end{tabular}
\end{table}

\textbf{Budget-matched analogues of recent methods.}
\label{sec:recent-sota}
\emph{These comparisons match search mechanisms and budgets, not published end-to-end results.} Five recent methods train an RL policy per candidate, using roughly four orders of magnitude more compute than our proxy \citep{Qiu2024,Song2025,RoboMoRe2025,Qiu2025,Strgar2026}. We therefore compare their search mechanisms under our genome, fitness, and budget. RoboGrammar's Graph Heuristic Search \citep{Zhao2020} and GLSO's latent-space BO \citep{Hu2022} are included similarly.

\emph{Coarse-to-fine is the strongest recent-method analogue and ties \panini.} The RoboMoRe analogue achieves $0.383\pm0.095$. Graph Heuristic Search, diversity-archive ES, and an LLM-conditioned proposer trail before correction ($p=0.014$--$0.033$), but not afterward ($p\ge0.77$). Latent-space BO is behind before correction ($p<10^{-3}$) and borderline afterward ($p=0.08$). Controller fine-tuning and archive criterion-rotation are significantly behind ($p\le0.004$).

\textbf{Budget dependence and CMA-ES implementation.}
\label{sec:budget-sweep}
\emph{Larger budgets do not resolve the tie with strong optimizers.} At budgets 60, 120, and 200 ($n=24$), \panini retains the best mean rank and fitness (Friedman $p<0.001$), but separates only from the two weakest references, never from GGE, PSO, or IPOP-CMA-ES (Nemenyi $p\ge0.85$; Figure~\ref{fig:budget}).

\emph{Standard IPOP-CMA-ES is the appropriate CMA-ES baseline.} It beats simplified CMA-ES on $19/24$ seeds ($p=1.2\times10^{-4}$). We consequently retain the simplified implementation only as a weak reference.

\textbf{Re-scoring final designs with RL-trained controllers.}
\label{sec:rl-rescoring}
Because proxy fitness weakly predicts trained performance, we re-scored final bodies using MLP controllers trained by antithetic evolution strategies in MuJoCo MJX under the same reward (Appendix~\ref{app:proxy-rl}).

\emph{RL training substantially changes rankings, but the earlier leading result comes from one repeated design.} An earlier collection contains 672 bodies from 28 methods and 24 seeds, predating \panini and GGE-v2 (Table~\ref{tab:rl28}). RL rankings differ significantly ($\chi^2=73.37$, $p=3.7\times10^{-6}$). Coordinate descent leads, but returns the same body on all 24 seeds, whereas every other method returns at least 23 distinct bodies. Its fitness spread therefore reflects controller-training noise, not varied design discovery. Across methods, proxy and RL ranks are weakly related ($\rho=0.17$, $p=0.39$; Figure~\ref{fig:rl-rescoring}).

\emph{The \panini-specific RL evaluation confirms competitiveness, not superiority.} This separate evaluation covers 144 bodies from six methods and 24 seeds. All converge, and rankings do not differ significantly ($\chi^2=2.95$, $p=0.71$). \panini ranks 2nd ($3.29$; $1.476\pm0.697$), behind PSO ($3.12$; $1.384\pm0.517$) and ahead of GGE ($3.33$), CEM ($3.62$), IPOP-CMA-ES ($3.75$), and simplified CMA-ES ($3.88$; Table~\ref{tab:rl6}, Figure~\ref{fig:rl-rescoring-panini}).

\begin{table}[h]
\caption{Earlier RL re-scoring: 672 bodies from 28 methods and 24 seeds, collected before \panini. Proxy ranks come from our re-run because the original output field contains NaN values. Nemenyi $p$-values compare with GGE. The separate \panini evaluation appears in Table~\ref{tab:rl6}.}
\label{tab:rl28}
\centering
\small
\begin{tabular}{lcccc}
\toprule
Method & RL mean rank & RL fitness & Proxy mean rank & Nemenyi $p$ \\
\midrule
Coordinate descent (1 distinct design) & 5.17 & $2.690\pm0.355$ & 19.88 & 0.0004 \\
IPOP-CMA-ES & 10.08 & $2.256\pm0.517$ & 11.46 & 0.505 \\
Novelty search & 11.08 & $2.121\pm0.638$ & 10.96 & 0.825 \\
Random search & 12.62 & $2.036\pm0.629$ & 17.94 & 0.996 \\
\multicolumn{5}{c}{\emph{(19 further methods: RL ranks 12.75--16.29, all $p\ge0.997$)}} \\
Grammar-guided evolution (GGE) & 16.75 & $1.642\pm0.924$ & 16.92 & --- \\
\multicolumn{5}{c}{\emph{(3 further methods: RL ranks 16.92--18.08, all $p=1.000$)}} \\
Nelder-Mead & 20.08 & $1.379\pm0.607$ & 20.54 & 1.000 \\
\bottomrule
\end{tabular}
\end{table}

\textbf{Feasibility gate versus task optimality.}
\label{sec:results-gate}
\emph{The gate reduces physics evaluations needed to reach feasibility, but does not improve best fitness.} Blind sampling needs $11.8\pm10.6$ genome draws to find a feasible body under the calibrated specification (24 seeds). The gate screens rejections before simulation, requiring one physics evaluation by construction.

Under 40 physics evaluations with free rejections, best fitness is $0.334\pm0.230$ with the gate versus $0.339\pm0.115$ without it ($p=0.91$; Table~\ref{tab:gate}). The gate wins 15/24 seeds; its larger variance mainly comes from one seed with no positive-fitness body. An earlier significant gate disadvantage does not reproduce after the geometry fix and is withdrawn.

\begin{table}[h]
\caption{Gate ablation: 40 physics evaluations, 24 seeds, and corrected geometry. Paired $t$-tests compare with no gate. Per-seed results: Appendix~\ref{app:rawdata}.}
\label{tab:gate}
\centering
\small
\begin{tabular}{lccc}
\toprule
& Best fitness & Wins vs.\ no gate & $p$ \\
\midrule
No gate & $0.339\pm0.115$ & --- & --- \\
Geometric gate & $0.334\pm0.230$ & 15/24 & $0.91$ \\
Geometric $+$ gait pre-check & $0.269\pm0.242$ & 9/24 & $0.20$ \\
\bottomrule
\end{tabular}
\end{table}

\emph{The evaluation saving grows as feasible designs become rarer.} Ungated sampling cost rises from 1.8 draws at $p=0.46$ to 124 at $p=0.0035$, while gated physics cost stays at one. Adding a gait pre-check does not help ($p=0.055$ versus the geometric gate). GGE reaches feasible genomes more often than blind sampling, significantly at three of four levels; the hardest remains unresolved ($7.8\%$ versus $0.6\%$, $p=0.19$, $n=8$; Appendix~\ref{sec:gait-correlated-gate}--\ref{sec:complex-env}).

\textbf{What the gate predicts.}
\emph{The gate enforces design requirements, not simulator stability.} Among 200 random bodies under each of two specifications, the five geometric checks rejected 27 and 180, yet all 400 loaded and stepped successfully with actuators and finite state. Its savings therefore concern reaching a chosen design target, not avoiding simulator crashes (Table~\ref{tab:gate-scaling}).

\textbf{Validity versus 14 alternative strategies.}
\label{sec:results-validity}
\emph{Grammar-guided mutation preserves structural validity, but three alternatives also reach $100\%$.} Across 400 trials per condition, re-initialization uses the grammar itself, mild rescaling barely changes the body, and range-constrained resampling edits an already coordinated parent. Other alternatives degrade substantially, with three reaching $0\%$ (Table~\ref{tab:validity21}, Appendix~\ref{app:validity-table}).

\emph{Higher validity does not necessarily produce higher task fitness.} Careful hand-coded duplication and deletion achieve only $324/400$ and $243/400$ valid bodies. However, exploratory raw-XML editing is less valid ($58.8\%$ versus $100\%$) but directionally ahead in best fitness ($0.769$ versus $0.351$, $p=0.23$, $n=8$; Appendix~\ref{sec:validity-vs-fitness}).

\textbf{Legality does not guarantee locomotion.}
\label{sec:stricter-validity-main}
\emph{The grammar's structural-validity advantage does not extend to locomotion against constrained resampling.} Under a $\ge2$\,cm displacement criterion, grammar mutation and range-constrained resampling tie: $370/400$ versus $377/400$ passes (Fisher $p=0.39$). Mild rescaling and re-initialization also tie, while disruptive alternatives fall as low as $0/400$ (Appendix~\ref{sec:stricter-validity}).

Thus, a well-formed body is not necessarily capable of locomotion. Joint cross-limb feasibility is a separate question.

\textbf{Coupled, cross-limb feasibility.}
\label{sec:coupled-feasibility}
The grammar spaces limbs evenly around the torso, allowing mutation jitter of $\pm0.3$ radians. Independent per-field ranges cannot enforce this joint relationship. We isolate its effect by keeping all other properties fixed and sampling mount angles independently, then measuring support-polygon area on the compiled model.

\emph{Coordinated placement outperforms independent mount angles at every tested threshold.} Pass rates are $23.5\%$ versus $12.3\%$, $12.0\%$ versus $5.3\%$, and $6.0\%$ versus $2.5\%$ ($p<10^{-4}$, $0.001$, and $0.023$; Table~\ref{tab:coupled-feasibility}). Range-constrained resampling ties at the easier levels and leads at the hardest ($10.3\%$ versus $6.0\%$, uncorrected $p=0.038$), but starts from grammar-coordinated parents. The cause of its hardest-level advantage remains unresolved.

\textbf{Grammar formalism or coordinated layout?}
\emph{Coordination drives the advantage; the grammar formalism is not uniquely required.} A hand-coded generator with the same layout and geometry, but no rule objects, roles, or arbitration, ties the grammar at every level: 87 versus 94 passes, 38 versus 48, and 32 versus 24 out of 400 ($p=0.61$, $0.30$, $0.33$). Randomizing its mount angles reduces passes to 42, 18, and 13 ($p<10^{-4}$, $0.009$, $0.006$).

The grammar's contribution is expressing that coordination as reusable rules across catalogues and domains, rather than as a one-off generator.

\textbf{Construction in a very-low-feasibility space.}
\label{sec:hard-design-space}
To remove the benefit of coordinated parents, we compare construction from scratch. Flat designs expose $8n_\ell$ independent limb parameters; grammar designs use the same four continuous parameters as elsewhere. Joint requirements cover support area, stability, angular gaps, tip separation, and mass. One scalar tightens the thresholds, calibrated to pooled pass rates across samplers.

\emph{Grammar construction finds feasible bodies where flat sampling and budget-limited flat CMA-ES find none.} With eight limbs, flat sampling finds no feasible body in 6000 draws at any level, while grammar construction passes $13.1\%$--$1.0\%$. At 200 evaluations and 20 seeds, flat CMA-ES finds a feasible body in $0/20$ seeds despite a continuous constraint-violation score. Grammar-CMA-ES succeeds in $20/20$, with $27$--$31\%$ feasible evaluations. Ten-limb tests show the same pattern (Table~\ref{tab:hard-space}, Appendix~\ref{app:validity-hard}).

This advantage is over generic flat parameterization, not hand-coded coordination. It also does not imply failure of flat CMA-ES at arbitrarily larger budgets.

\label{sec:validity-hard}
\emph{Stricter joint constraints separate mutation methods that previously tied.} Repeating the 15-condition protocol on feasible parents, the grammar preserves feasibility in $285/300$ trials versus $166/300$ for range-constrained resampling at the $8\%$ level; at $0.3\%$, the counts are 169 versus 129 (Fisher $p\le1.4\times10^{-3}$).

\emph{The grammar also produces more genuinely new feasible bodies and can change topology.} Near-matching alternatives mostly leave the parent unchanged: mild rescaling moves a leg tip in $15\%$ of trials by $0.4$\,mm, versus $82\%$ and $15.3$\,mm for the grammar. Counting only genuinely new feasible bodies, the grammar leads every single-parent operator, reaching $77\%$ versus at most $18\%$ at the $8\%$ level. Range edits and clipping cannot change topology and succeed in $0/20$ search seeds; grammar search can add limbs and succeeds in $20/20$, with median 10 evaluations.

\textbf{Four further checks.}
\label{sec:objections}
Additional tests use eight limbs and all four difficulty levels (Table~\ref{tab:objections}, Appendix~\ref{app:objections}).

\emph{Repeated mutations favor the grammar over clipping.} One mutation nearly ties, but after ten, feasibility is $86.7\%$ for the grammar versus $30.0\%$ for clipping at the easiest level (Fisher $p<0.01$ throughout). In material-minimization search, the grammar beats the best of five flat variants on $20/20$ seeds at every level (Wilcoxon $p=1.9\times10^{-6}$).

\emph{The advantage persists with a larger grammar genome.} Per-field boxes around feasible designs remain $19$--$35\%$ feasible in the five-parameter representation but at most $0.5\%$ in flat 64-dimensional space, and $0\%$ after rotation. A 12-parameter grammar genome still lets CMA-ES succeed in 6--20 of 20 seeds, versus $0/20$ for flat CMA-ES.

\emph{Hand-coded editors match only when they implement the layout rule.} Inserting a limb into the largest gap alone never succeeds.

\emph{The gate alone does not explain the advantage.} Flat generation finds no feasible body in 6000 draws, with or without the gate.

\section{Searching for a Grammar-Native Search Advantage}
\label{sec:edge}

Can grammar structure improve search dynamics, rather than only define the shared genome? We tested heterogeneous per-limb spaces, fitness noise, subtree crossover, quality-diversity selection, motif reuse, complexity tempering, production repair, transfer, and low-budget regimes using the same harness and budget accounting (Appendix~\ref{app:edge}).

\emph{Most grammar-native search variants do not outperform strong generic optimizers; two perform worse.} Except for the curriculum below, none establishes an advantage. In the larger heterogeneous space, \panini falls to 3rd of 4 methods ($n=10$), contrary to expectations that grammar benefits grow with design-space size \citep{Zhao2020,Hu2022}. A prior transferred from 2--5 to 6--8 legs also loses to cold-start IPOP-CMA-ES and PSO (Nemenyi $p=0.012$, $0.038$). Compositional structure can therefore hinder search.

\paragraph{The One Positive Result: A Curriculum over Grammar Depth.}
\label{sec:curriculum}

The curriculum initially restricts leg and segment counts, $n_\ell$ and $k$, to 35\% of their ranges, then expands their upper bounds linearly over the budget \citep{Bengio2009}. Initially, this permits at most 4 legs and mostly single-segment limbs. All other \panini components remain unchanged. The schedule was selected from an unreplicated $n=10$ screen, where all tested schedules exceeded plain \panini ($0.460$--$0.471$ versus $0.396$).

\begin{table}[h]
\caption{Curriculum versus plain \panini and seven references. The curriculum ranks first in all four runs. Nemenyi $p$-values compare with curriculum and are not corrected across variants. The $n=48$ runs include the original 24 seeds plus 24 new seeds.}
\label{tab:curric}
\centering
\footnotesize
\begin{tabular}{lcccc}
\toprule
& \multicolumn{2}{c}{$n=24$} & \multicolumn{2}{c}{$n=48$} \\
& $b=60$ & $b=120$ & $b=60$ & $b=120$ \\
\midrule
Curriculum / plain \panini fitness & $0.459/0.411$ & $0.508/0.455$ & $0.449/0.388$ & $0.490/0.429$ \\
Friedman $p$ & $1.1\times10^{-4}$ & $4.5\times10^{-5}$ & $2.7\times10^{-5}$ & $4.5\times10^{-6}$ \\
\midrule
vs.\ IPOP-CMA-ES & 0.486 & 0.134 & 0.027 & 0.051 \\
vs.\ PSO & 0.486 & 0.579 & 0.004 & 0.046 \\
vs.\ Cross-entropy method & 0.346 & 0.194 & 0.358 & 0.034 \\
vs.\ Coarse-to-fine & 0.285 & 0.134 & 0.0009 & 0.016 \\
vs.\ TuRBO & 0.205 & 0.057 & 0.017 & 0.0004 \\
vs.\ LHS + refinement & 0.023 & 0.004 & 0.034 & 0.001 \\
vs.\ CMA-ES (simplified) & $<10^{-4}$ & $<10^{-4}$ & $<10^{-4}$ & $<10^{-4}$ \\
vs.\ plain \panini & 0.579 & 0.815 & 0.262 & 0.175 \\
\bottomrule
\end{tabular}
\end{table}

\emph{The curriculum leads numerically at both budgets, with stronger evidence at the larger sample size.} At $n=24$, it leads by about $0.05$ over plain \panini, but separates only from the two weakest references. At $n=48$, it is nominally ahead of six of seven references at budget 60 and five at budget 120; IPOP-CMA-ES is borderline at the latter ($p=0.051$).

\emph{The improvement is modest and sensitive to statistical correction.} Direct paired comparisons with plain \panini show gains of $+0.060$ ($p=0.003$) and $+0.061$ ($p=0.001$), but Nemenyi comparisons remain non-significant. Three caveats apply: correction across the tested variants leaves only simplified CMA-ES significant at both budgets; the larger sample includes the original seeds rather than independently replicating them; and early preference for small bodies is confounded with gradual expansion.

\emph{The grammar's clearest contributions are structural, not a general optimization advantage.} Strong generic optimizers remain difficult to beat on this genome. The curriculum offers a modest, correction-sensitive improvement; the strongest findings remain compact representation, coordinated construction, and preservation of joint constraints.

\section{Conclusion}
\label{sec:conclusion}

A typed case-role production grammar with inheritance and specificity-based arbitration admits a checkable geometric semantics: rule composition matches $SE(3)$ composition wherever inheritance is inactive, with a small, measured defect elsewhere. The grammar is compact, and its conflict resolution is recoverable as geometric containment, robust to regularization, embedding geometry, and perception noise.

As a search space the grammar is shared by every optimizer we tested, and over it \panini is tied with the strongest generic methods -- best mean rank of 30, but statistically indistinguishable from IPOP-CMA-ES, CEM, PSO, and TuRBO -- a tie that RL re-scoring of 672 final designs does not change. Further attempts to give the grammar a role in the search dynamics produced ties, one significant loss, and one small, marginal positive effect (the depth curriculum). We take this consistency as the result: on this task, well-tuned generic optimizers are hard to beat however the grammar participates in search.

The grammar's contribution is instead structural. It is legal by construction where 14 alternatives are not, and its coordinated limb placement keeps edits jointly feasible where range-constrained resampling does not; in a very-low-feasibility space, flat parameterization finds no feasible design in 6,000 draws, while only the grammar reaches designs with a different limb count. These advantages hold over flat parameterizations, not over a hand-written generator encoding the same coordination, and have measured limits: legality does not imply locomotion, and the geometric gate encodes a design specification rather than predicting simulator failure.

\bibliographystyle{unsrt}  
\bibliography{references}  

\newpage
\appendix

\section{Full Production Rule Catalogues}
\label{app:rules}

\subsection{Manipulation Grammar (Eight Rules)}
\begin{itemize}
\item \textbf{Approach (generic).} Requires Agent, Patient. Emits \emph{aligned}. Parameters: approach height (default 0.10\,m). Specificity 1. Exports approach height.
\item \textbf{Approach (tool variant).} Requires Agent, Patient, Instrument. Emits \emph{aligned}. Parameters: approach height (default 0.08\,m). Specificity 2 (overrides the generic variant whenever an Instrument is bound). Exports approach height.
\item \textbf{Grasp.} Requires Agent, Patient. Consumes \emph{aligned}. Emits \emph{holding}. Parameters: grip force (default 5\,N). Specificity 1. Imports approach height.
\item \textbf{Lift.} Requires Agent, Patient. Consumes \emph{holding}. Parameters: lift height (default 0.15\,m). Specificity 1.
\item \textbf{Transport.} Requires Agent, Patient, Goal. Consumes \emph{holding}. Parameters: transport speed (default 0.2\,m/s). Specificity 1.
\item \textbf{Place.} Requires Agent, Patient, Location. Consumes \emph{holding}. Emits \emph{placed}. Parameters: release height (default 0.02\,m). Specificity 1.
\item \textbf{RotateTool.} Requires Agent, Instrument. Consumes \emph{holding}. Parameters: rotation angle (default $90^\circ$). Specificity 1.
\item \textbf{Retract.} Requires Agent, Source. Consumes \emph{placed} or \emph{holding}. Parameters: retract distance (default 0.20\,m). Specificity 1.
\end{itemize}

\subsection{Morphology Grammar (Seven Rules)}
\begin{itemize}
\item \textbf{AddRoot.} Requires Root. Emits \emph{has\_root}. Parameters: torso length/width/height (0.30/0.15/0.10\,m), torso mass (3.0\,kg). Specificity 1.
\item \textbf{AddLimb (generic).} Requires Root, Limb. Consumes \emph{has\_root}. Emits \emph{has\_limb}. Parameters: mount angle, segment length (0.12\,m default), segment radius (0.02\,m default). Specificity 1. Exports segment length, radius.
\item \textbf{AddLimb (extended-reach).} Requires Root, Limb. Consumes \emph{has\_root}, \emph{needs\_reach}. Emits \emph{has\_limb}. Parameters: mount angle, segment length (0.18\,m), segment radius (0.018\,m). Specificity 2. Exports segment length, radius.
\item \textbf{Articulate.} Requires Limb, Joint. Consumes \emph{has\_limb} or prior \emph{articulated}. Emits \emph{articulated}. Parameters: joint range ($\pm90^\circ$ default), joint type (revolute). Specificity 1.
\item \textbf{Extend.} Requires Limb, Segment. Consumes \emph{articulated}. Emits \emph{extended}. Specificity 1. Imports segment length, radius.
\item \textbf{Terminate.} Requires Limb, Terminal. Consumes \emph{extended} or \emph{has\_limb}. Emits \emph{terminated}. Parameters: foot radius (0.015\,m), foot mass (0.05\,kg). Specificity 1.
\item \textbf{Symmetrize.} Requires Limb. Consumes \emph{terminated}. Optional. Parameters: mirror axis (sagittal plane default). Specificity 1.
\end{itemize}

\section{Implementation Notes}
\label{app:pitfalls}

This appendix documents implementation decisions and validation checks that materially affect the numbers reported in the main text.

Algorithm~\ref{alg:derive} gives the derivation procedure Section~\ref{sec:method} describes in prose, exactly.

\begin{algorithm}[h]
\caption{Grammar Derivation}
\label{alg:derive}
\KwIn{Plan $P = (\text{op}_1, F_1), \ldots, (\text{op}_n, F_n)$; rule catalogue $\mathcal{G}$}
\KwOut{Derivation $T = (a_1, \ldots, a_n)$, each $a_i$ an applied rule with resolved parameters}
Initialize marker set $M \leftarrow \emptyset$; inherited-parameter map $I \leftarrow \emptyset$\;
\For{$i \leftarrow 1$ \KwTo $n$}{
  $C \leftarrow \{r \in \mathcal{G} : r.\text{op} = \text{op}_i,\ r.\text{req} \subseteq \text{roles}(F_i),\ r.\text{cond}(F_i, M) = \text{true}\}$\;
  \If{$C = \emptyset$}{\textbf{fail}: no applicable rule for step $i$\;}
  $r^\star \leftarrow \arg\max_{r \in C} s(r)$, ties broken by declaration order\;
  $\theta \leftarrow r^\star.\text{params}$\;
  \For{$k \in r^\star.\text{imports} \cap \text{keys}(I)$}{
    $\theta[k] \leftarrow I[k]$ \tcp*{inherited value overrides own default}
  }
  \If{step $i$ supplies an explicit override for $k$}{$\theta[k] \leftarrow$ override value\;}
  $a_i \leftarrow$ instantiate $r^\star$ with resolved parameters $\theta$\;
  $M \leftarrow (M \setminus r^\star.\text{consume}) \cup r^\star.\text{emit}$\;
  \For{$k \in r^\star.\text{exports}$}{$I[k] \leftarrow \theta[k]$ \tcp*{write forward for later steps}}
}
\Return{$T = (a_1, \ldots, a_n)$}\;
\end{algorithm}

\textbf{The floating-base torso and integrator stability.} Every compiled body's torso -- the one part meant to move freely under gravity -- is attached to the world by an explicit six-degree-of-freedom free joint. Forward center-of-mass displacement therefore reflects genuine translation of the whole body, rather than the angle of an actuated hip hinge measured against a kinematically fixed base. Allowing the torso to fall and swing under real dynamics makes roughly half of sampled bodies numerically unstable under a naive explicit integrator, since actuated hinge joints are otherwise asked to support a body actually falling under gravity with no damping or armature. We stabilize this with per-joint damping ($0.3$) and armature ($0.01$), a smaller timestep ($0.002$\,s), and an implicit integrator: standard stabilizers for actuated hinges under contact, verified not to change which designs the geometric feasibility checks accept. On a smoke test of 100 freshly sampled 8-limb bodies, this configuration shows 0 numerical blowups and 0 degenerate (zero-actuator) bodies, with displacement over a 2-second horizon ranging $-0.34$ to $+0.88$\,m (mean $0.13$\,m) -- a physically plausible range for a body of this scale. One residual imperfection remains and is disclosed rather than papered over. With the torso genuinely falling and limbs genuinely swinging under gravity, a folded limb settles into geometric interpenetration with the torso at the contact point. This is measured directly over $n=200$ sampled bodies at median $1.42$\,cm, $95$th percentile $3.42$\,cm, and worst-observed $5.50$\,cm, without causing a numerical blowup in any of the 200 bodies: the default soft-contact solver is tolerating real penetration rather than fully resolving it. This distinction matters for how ``valid'' should be read throughout this paper. Syntactic validity (a well-formed derivation), geometric validity (passing the feasibility checks on the forward-kinematic tree, before any physics step), and physical validity (behaving acceptably once simulated, including bounded self-contact penetration) are three different properties, and every validity number in Sections~\ref{sec:results-validity}--\ref{sec:objections} is of the first two kinds only.

Four further checks address this directly. First, does a stricter contact solver change the picture? Re-running the same 200 bodies with a tightened contact solver and more solver iterations drops median penetration to $0.25$\,cm, but raises the $95$th percentile to $4.66$\,cm and the worst case to $7.84$\,cm, with 0 of 200 bodies failing to load or step under the tighter tolerances. Second, does this change the fitness numbers the rest of the paper reports? Re-evaluating the same genomes' same compiled geometry under the stricter settings gives mean fitness $+0.0313\pm0.1183$ against $+0.0244\pm0.0380$ under the default settings used throughout: a similar point estimate with visibly higher variance, not a directional bias. The main search and validity results are therefore not an artifact of under-resolved contact. Third, can an explicit penetration threshold with a repair stage bring this under control at generation time, rather than only being measured post hoc? At a $2.0$\,cm threshold, a bounded repair loop (shrink limb/torso radius by $15\%$ per attempt, up to 5 attempts, re-deriving through the normal grammar path each time) accepts $88/100$ bodies as drawn, repairs a further $7/100$ (mean $2.3$ attempts), and rejects $5/100$ as unfixable. This is a concrete, measured alternative to tolerating unbounded penetration, offered as a deployment option rather than wired into the main search loop by default, since doing so would change the generator itself for every other reported number. Fourth: does wiring that repair/rejection stage into GGE's search loop change its result? Under the corrected geometry, re-running GGE (budget $60$, $n=8$ seeds) with every candidate checked against the $2.0$\,cm threshold before evaluation, repaired or rejected as above, accepts $396/480$ genomes as drawn, repairs $7$, and rejects $77$ ($16.0\%$). Best fitness falls from $0.353\pm0.104$ without the stage to $0.249\pm0.116$ with it on the same seeds (paired $p=0.10$). An earlier version of this check reported no change against a pre-fix reference; that comparison is superseded. The stage is therefore a real trade-off between bounded penetration and fitness, and it stays disabled for every other number in this paper.

\textbf{Actuator gain and fitness validity.} The actuator gain is calibrated so that center-of-mass displacement under the fixed sinusoidal controller varies with leg count, together with an implausible-displacement guard; a gain too weak to drive real locomotion would give identical displacement across different leg counts, which is physically impossible for a genuine gait, so this check is run on every change to the controller or genome space.

\textbf{Budget accounting in the gate ablation.} The gate ablation (Section~\ref{sec:results-gate}) counts only completed physics evaluations against the shared budget, not every genome attempted; counting free gate rejections against the same budget would throttle gate-active search to a far smaller attempt count than gate-inactive search and confound the comparison.

\textbf{Severity of the adversarial mutation baseline.} The adversarial mixture in Section~\ref{sec:setup-validity} combines collapse-to-zero, sign-flip, and five-to-fifty-fold blow-up perturbations rather than mild rescaling alone, so that this baseline is not a low bar that produces a misleadingly high validity rate.

\textbf{Decoupling feature count from specificity.} The specificity hierarchy in Section~\ref{sec:boxes} is built across two independent rule families so that a general rule carries a feature its override does not share, decoupling feature overlap from true specificity; a hierarchy where every override is a literal feature superset of its general counterpart would let a Euclidean-nearest-neighbor control with no notion of containment score perfectly by coincidence.

\textbf{Noise model for the perception-noise test.} The perception-noise sweep (Section~\ref{sec:boxes}) models noise as a discrete bit-flip on the actual binary features used to compute the candidate set, since the arbitration decision does not depend on continuous state once that candidate set is fixed -- a continuous perturbation of state alone would not exercise the decision at all.

\textbf{Budget enforcement in IPOP-CMA-ES.} The external CMA-ES implementation's population-based stopping criterion is checked only between generations, which can overshoot a fixed evaluation budget (up to 49 evaluations for a budget of 40 in this genome space); an effective cap below the nominal budget, verified across four budget sizes, keeps every method's accounting comparable.

\textbf{Unit normalization in SPSA and Nelder-Mead.} Both methods operate on a normalized $[0,1]^5$ representation of the genome, converting to real units only at evaluation time, verified against a synthetic objective with a known optimum. This matters because the genome's real-unit dimensions span a factor of $150$ between the largest and smallest range; operating directly in real units, or mishandling SPSA's internal scale factor, can inflate a method's effective step size by up to $36\times$ in one dimension relative to the others.

\textbf{Structure-aware inheritance check.} The per-limb inheritance check in Appendix~\ref{sec:inheritance-ablation} compares each limb's first-segment radius and length in the compiled body with the genome's, limb by limb, rather than counting matching elements anywhere in the body (which coincidental matches can satisfy) or checking whether the value appears anywhere as a substring -- a substring check passes unconditionally from the first limb's explicit override alone and cannot detect whether any other limb actually reflects it.

\textbf{Per-evaluation feasibility measurement.} The shrinking-feasible-region search comparison (Appendix~\ref{sec:complex-env}) measures validity directly at every evaluation call, rather than reading it off a method's running best-fitness-so-far trace, which reads as feasible at every entry once any feasible genome has been found, regardless of whether a given draw was itself feasible, and so inflates whichever method happens to find a feasible point earliest. Measuring per-evaluation feasibility directly removes this confound and is the methodology behind every number in Appendix~\ref{sec:complex-env}.

\section{Extended Results and Diagnostics}
\label{app:extended}

This appendix gives the full account, tables, and figures for results summarized briefly in Section~\ref{sec:results}.

\subsection{Inheritance Ablation: Does It Actually Reach Every Limb?}
\label{sec:inheritance-ablation}

The compactness result (Section~\ref{sec:compactness}) is a rule-\emph{counting} argument: inheritance reduces the number of distinct (operation, parameter-tuple) rules needed when every limb shares the same segment geometry. It does not, by itself, establish that inheritance is \emph{executing} as described -- that every limb's compiled geometry reflects the genome's current segment radius, not only the first limb's. We test this with two checks: an ablation (every rule's inheritance-export set forced empty) measured against mutation validity, and a structure-aware trace checking, limb by limb, whether each limb's first segment in the compiled body carries the genome's segment radius and length (to the compiler's 3-decimal precision).

Validity is unaffected by the ablation, as a mechanism-level argument predicts ($100/100$ valid with inheritance active, $100/100$ disabled, $n=100$): MuJoCo's load-and-step check has no way to know whether a geometry's intended parameter was honored, only whether the resulting body is physically well-formed, which typing/legality gating (not inheritance) is responsible for.

The second check depends on which value takes precedence when a rule's own declared default and an inherited value share the same key. We give an inherited value precedence over a rule's own default for the same key, with an explicit per-step override, used only by the first limb, taking final precedence over both. Table~\ref{tab:inherit-fixed} reports the result. With inheritance active, every limb of every body carries the genome's mutated radius and length. With inheritance disabled, limbs after the first fall back to the rule's own defaults: no non-first limb matches the genome's length, and the $8/415$ radius matches are genomes whose radius happens to round to the default. Matches are counted per limb rather than anywhere in the body, so coincidental matches on later segments are excluded.

\begin{table}[h]
\caption{Inheritance ablation, $n=100$ mutated bodies (515 limbs; seed 0, two mutations each.}
\label{tab:inherit-fixed}
\centering
\small
\begin{tabular}{lccc}
\toprule
& Valid & Bodies, all limbs match & Non-first limbs: radius / length \\
\midrule
Inheritance active & $100/100$ & $\mathbf{100/100}$ & $415/415$ / $415/415$ \\
Inheritance disabled & $100/100$ & $\mathbf{0/100}$ & $8/415$ / $0/415$ \\
\bottomrule
\end{tabular}
\end{table}

This parameter-precedence choice does not affect the rule-counting compactness claim, a static property of the catalogue, nor the validity claim (Section~\ref{sec:results-validity}). It does mean that the choice matters for any result that depends on segment-radius or segment-length mutation reaching every limb rather than only the first: under the opposite precedence (a rule's own default always wins), such a mutation would only ever change the first limb's geometry, a strictly weaker effective mutation operator for two of the five genome parameters. Every fitness-dependent result in this paper uses the inherited-value-precedence order described above throughout, so this is a documented design choice rather than a source of inconsistency across experiments.

\subsection{Compositional-Generalization Anomaly: Full Diagnosis}
\label{sec:compgen-diagnostic}

The most likely innocent explanation for held-out error being \emph{lower} than in-distribution error (Table~\ref{tab:compgen}) is a length confound: held-out derivations might simply be shorter on average, giving twist-composition error less opportunity to accumulate. We test this across $n=5$ seeds (600 training derivations, 100 in-distribution and 100 held-out test derivations per seed, 60 epochs each) and bucket every evaluated derivation by its exact length.

Across seeds, the anomaly is stable: in-distribution error is $0.0612\pm0.0120$ and held-out error is $0.0124\pm0.0018$, held-out lower in all 5 of 5 seeds. Bucketing by length rules out the simple length confound, but in a way that itself needs explaining: the two splits barely overlap in length (in-distribution derivations are lengths 3, 4, 5; held-out are lengths 4 and 8), and at the one length they share, length 4, in-distribution error ($0.1705$, $n=178$) is over $14\times$ \emph{higher} than held-out error ($0.0118$, $n=244$) -- the opposite of a simple "held-out is shorter" story.

Tracing this length-4 split resolves it completely. Of the three in-distribution training templates, exactly one has length 4 and includes a tool-rotation step, and every held-out template of length 4 contains no such step. Splitting every evaluated derivation by whether it contains a rotation step, rather than by length, shows why this matters: derivations containing it have twist MSE $0.1806$ ($n=27$) versus $0.0010$ ($n=73$) for derivations without one -- a $180\times$ gap. Rotation is the only transformation whose twist depends on a per-scene, randomly sampled tool axis rather than a task-geometry-determined translation target, making it intrinsically harder to fit regardless of where it falls in a derivation. Because this one hard operation appears only in a training template and never in any held-out template, the in-distribution average is inflated by a task type the held-out split never has to handle.

The original anomaly is explained, not merely narrowed: it is an artifact of unequal task-type composition between the training and held-out template sets, specifically that the one rotation-based operation was placed exclusively on the training side. It does not reflect compositional generalization behaving backwards, and it does not bear on the morphology-domain result in Table~\ref{tab:compgen}, which uses a structurally different held-out split (leg and segment count, not primitive-rule identity) and shows the expected direction. A future version of this diagnostic should include a rotation-bearing template in the held-out set as well, so task difficulty is balanced across splits.

\subsection{Full Search Comparison}
\label{sec:search-full}

\begin{figure}[h]
\centering
\includegraphics[width=0.72\linewidth]{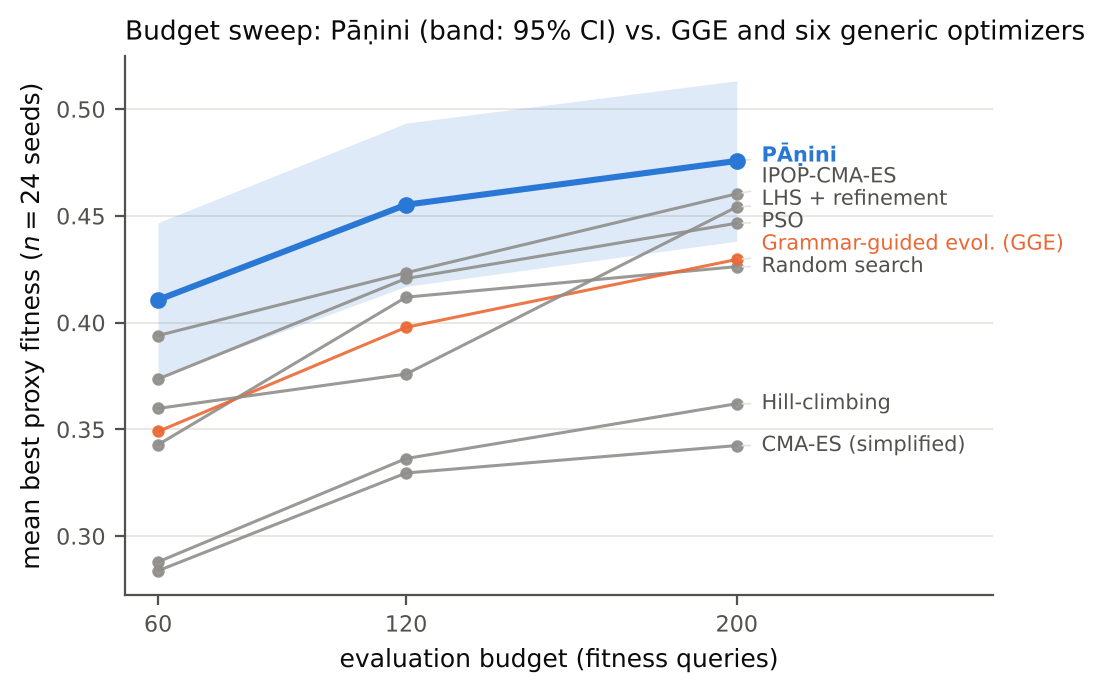}
\caption{Budget sweep, $n=24$ seeds, \panini against GGE and six reference methods (band: 95\% CI of the \panini mean). \panini has the best mean rank at every budget but is Nemenyi-significant only against the two weakest methods (simplified CMA-ES, hill-climbing), never against GGE or the strong cluster.}
\label{fig:budget}
\end{figure}

\begin{figure}[h]
\centering
\includegraphics[width=0.75\linewidth]{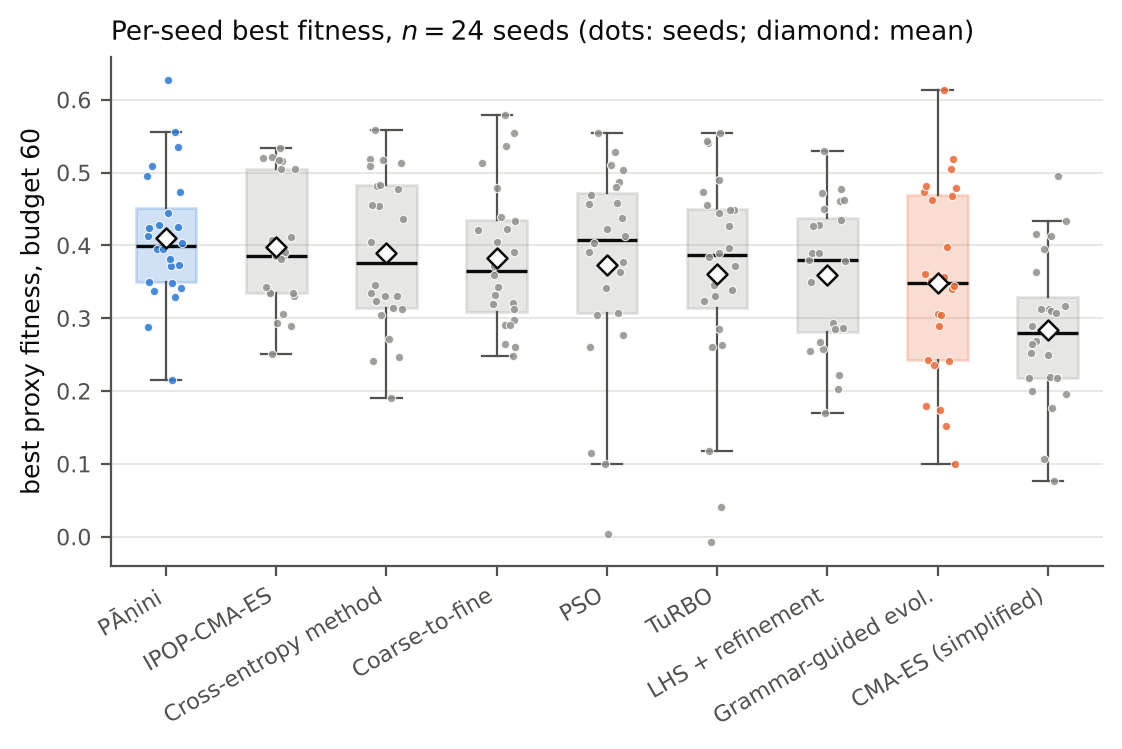}
\caption{Per-seed best fitness at budget 60, $n=24$, for \panini, GGE, and the reference cluster.}
\label{fig:boxplot}
\end{figure}

Table~\ref{tab:search30-full} gives all 30 methods of Table~\ref{tab:search30} (budget 60, $n=24$, post-fix geometry). Nemenyi $p$ is given against both grammar-based searchers. Figure~\ref{fig:budget} shows the \panini budget sweep of \S\ref{sec:budget-sweep}. At budget 200, NES, run in a separate file with the same seeds, scores $0.405\pm0.082$, below GGE. Figure~\ref{fig:boxplot} shows per-seed distributions; no aggregate is driven by a single seed. The paper's earlier 21-method and budget-sweep tables predate the geometry fix and are superseded.

\begin{table}[h]
\caption{All 30 search methods, budget 60, $n=24$ seeds, sorted by mean rank (Friedman $\chi^2=134.15$, $p=1.8e-15$). ``an.'' marks a budget-matched analogue of a recent method (\S\ref{sec:recent-sota}). Bold: Nemenyi $p<0.05$.}
\label{tab:search30-full}
\centering
\scriptsize
\begin{tabular}{lcccc}
\toprule
Method & Mean $\pm$ std. & Mean rank & Nemenyi $p$ vs.\ \panini & vs.\ GGE \\
\midrule
\textbf{\panini} & $0.411\pm0.088$ & 9.33 & --- & 1.000 \\
IPOP-CMA-ES & $0.397\pm0.086$ & 10.04 & 1.000 & 1.000 \\
Cross-entropy method & $0.390\pm0.104$ & 10.88 & 1.000 & 1.000 \\
PSO & $0.374\pm0.138$ & 11.19 & 1.000 & 1.000 \\
Coarse-to-fine (RoboMoRe an.) & $0.383\pm0.095$ & 12.04 & 1.000 & 1.000 \\
TuRBO & $0.361\pm0.144$ & 12.17 & 1.000 & 1.000 \\
GGE-v2 & $0.369\pm0.125$ & 12.58 & 1.000 & 1.000 \\
Augmented random search & $0.346\pm0.112$ & 13.25 & 1.000 & 1.000 \\
Grammar-guided evolution (GGE) & $0.349\pm0.130$ & 13.33 & 1.000 & --- \\
Diversity-archive ES (QD co-design an.) & $0.350\pm0.092$ & 13.56 & 0.999 & 1.000 \\
LHS + refinement & $0.360\pm0.097$ & 13.58 & 0.999 & 1.000 \\
Random search & $0.343\pm0.095$ & 14.52 & 0.973 & 1.000 \\
Graph heuristic search (RoboGrammar an.) & $0.326\pm0.131$ & 14.56 & 0.970 & 1.000 \\
SHADE & $0.341\pm0.101$ & 14.73 & 0.956 & 1.000 \\
Bayesian optimization & $0.349\pm0.112$ & 14.81 & 0.947 & 1.000 \\
Novelty search & $0.325\pm0.093$ & 14.92 & 0.935 & 1.000 \\
SPSA & $0.340\pm0.113$ & 14.96 & 0.929 & 1.000 \\
Differential evolution & $0.327\pm0.087$ & 15.27 & 0.878 & 1.000 \\
NES & $0.325\pm0.102$ & 15.69 & 0.781 & 1.000 \\
LLM proposer (RoboMorph/LASeR an.) & $0.329\pm0.129$ & 15.71 & 0.775 & 1.000 \\
$(1{+}1)$-ES & $0.291\pm0.165$ & 15.88 & 0.728 & 1.000 \\
MAP-Elites & $0.303\pm0.113$ & 17.71 & 0.195 & 0.998 \\
Simulated annealing & $0.297\pm0.104$ & 17.71 & 0.195 & 0.998 \\
Hill-climbing & $0.288\pm0.113$ & 18.17 & 0.119 & 0.989 \\
Latent-space BO (GLSO an.) & $0.296\pm0.093$ & 18.48 & 0.082 & 0.975 \\
CMA-ES (simplified) & $0.284\pm0.100$ & 18.98 & \textbf{0.043} & 0.926 \\
Coordinate descent & $0.282\pm0.000$ & 20.33 & \textbf{0.005} & 0.584 \\
Fine-tuned controller (Strgar an.) & $0.243\pm0.133$ & 20.58 & \textbf{0.004} & 0.502 \\
Archive criterion-rotation (Debate2Create an.) & $0.202\pm0.123$ & 23.58 & \textbf{$<10^{-4}$} & \textbf{0.018} \\
Nelder-Mead & $-0.054\pm0.315$ & 26.46 & \textbf{$<10^{-4}$} & \textbf{$<10^{-4}$} \\
\bottomrule
\end{tabular}
\end{table}

\subsection{Feasibility Gate: Gait-Correlated Check and Scaling}
\label{sec:gait-correlated-gate}

Every check in the geometric gate is static and blind to the locomotion metric. The third row of Table~\ref{tab:gate} adds one check that is correlated with gait: a short CPG rollout (one-tenth of the full horizon) that rejects a body moving less than $1$\,cm. Unlike the geometric checks, it costs $0.1$ of a physics evaluation per call against the same budget (40, $n=24$, same seeds). It is the lowest of the three conditions: $0.269\pm0.242$, against $0.334\pm0.230$ for the geometric gate alone (paired $t$ $p=0.055$, Wilcoxon $p=0.031$; it wins 7 of 24 seeds) and $0.339\pm0.115$ for no gate ($p=0.20$). At this budget the pre-check's cost outweighs whatever it gains by steering away from immobile bodies. All 72 best bodies (3 conditions $\times$ 24 seeds) compile, load, and step 1000 steps in MuJoCo with finite state.

\subsubsection{Does the Gate's Advantage Scale as the Feasible Region Shrinks?}
\label{sec:gate-scaling}

The tries-to-first-feasible figure in Section~\ref{sec:results-gate} is one operating point. We sweep the specification's reach threshold to four target pass rates and count the uniformly random genomes drawn before the first feasible one (gate off), against gate-on cost, which is $1$ by construction. The pre-fix reach thresholds are unreachable under the corrected geometry, so they were recalibrated as upper quantiles of reach over 4000 fresh genomes. Measured $p$ comes from an independent sample of 2000.

\begin{table}[h]
\caption{Feasibility-gate cost scaling across four difficulty levels (24 seeds per level for gate-off tries, post-fix geometry).}
\label{tab:gate-scaling}
\centering
\small
\begin{tabular}{lcccc}
\toprule
Target feasibility $p$ & $0.50$ & $0.20$ & $0.05$ & $0.01$ \\
\midrule
Required reach & $0.231$ & $0.367$ & $0.495$ & $0.561$ \\
Measured $p$ & $0.463$ & $0.177$ & $0.040$ & $0.0035$ \\
Predicted $1/p$ & $2.2$ & $5.7$ & $25.0$ & $285.7$ \\
Gate-off mean tries (std) & $1.8\ (1.1)$ & $3.8\ (2.7)$ & $21.2\ (19.5)$ & $124.4\ (130.7)$ \\
Gate-on cost (by construction) & $1.0$ & $1.0$ & $1.0$ & $1.0$ \\
\bottomrule
\end{tabular}
\end{table}

Gate-off cost grows with difficulty in the direction a $1/p$ law predicts, from under two tries to $124$, while gate-on cost stays at $1.0$. At the hardest level the measured $p$ (7 of 2000) falls below the $0.01$ target and is itself imprecise, so $1/p$ there is only a rough guide. This claim -- the feasibility \emph{gate} (check-before-committing) scaling with difficulty -- is a different mechanism from grammar-guided \emph{mutation locality} (Appendix~\ref{sec:complex-env}), and the two should not be conflated: the gate result holds at all four levels, and the mutation-locality result is significant at three of four and directionally ahead at the fourth.

\subsection{Grammar-Guided Search Under Shrinking Feasibility}
\label{sec:complex-env}

A separate question from gate scaling is whether grammar-guided \emph{mutation} has an advantage in a low-feasibility environment independent of the gate: once a search has found one feasible genome, does a small mutation land on another feasible genome more often than redrawing a fresh genome uniformly at random? If the feasible region is spatially contiguous, local mutation should increasingly outperform blind resampling as the feasible fraction shrinks.

\textbf{Measurement.} Feasible-eval-rate is measured by recording validity directly at every evaluation call, rather than reading it off a running \emph{best-so-far} trace (also noted in Appendix~\ref{app:pitfalls}): once any feasible genome is found, every later entry in such a trace reads as feasible regardless of that step's own draw, which would silently inflate whichever method happens to find a feasible point earliest. The numbers below use the per-evaluation measurement throughout.

We ran the $(\mu{+}\lambda)$ evolutionary search of Section~\ref{sec:results-search} and a blind-random-search baseline under the same gate, at the four recalibrated levels of Table~\ref{tab:gate-scaling}, together with two seeding variants. One injects a fresh random draw with probability $0.3$ per generation; the other refuses to switch to mutation until it holds a full population of feasible genomes. We measured the fraction of each method's budget that landed on a feasible genome (8 seeds, budget 40).

\begin{table}[h]
\caption{Feasible-evaluation rate (per-evaluation measurement), same gate, 8 seeds, budget 40, post-fix geometry. $p$: paired $t$-test on per-seed rates against blind random search.}
\label{tab:complex-env}
\centering
\small
\begin{tabular}{lcccc}
\toprule
Target feasibility $p$ (measured) & $0.50$ ($0.476$) & $0.20$ ($0.156$) & $0.05$ ($0.048$) & $0.01$ ($0.006$) \\
\midrule
Grammar-guided evolution & $80.0\%$ & $61.6\%$ & $28.7\%$ & $7.8\%$ \\
\quad + random restarts & $73.1\%$ & $41.2\%$ & $17.8\%$ & $3.8\%$ \\
\quad + feasible seeding & $72.8\%$ & $35.3\%$ & $5.3\%$ & $0.6\%$ \\
Blind random search & $47.8\%$ & $17.8\%$ & $2.8\%$ & $0.6\%$ \\
\midrule
$p$, evolution & $<10^{-4}$ & $0.003$ & $0.024$ & $0.19$ \\
$p$, + restarts & $<10^{-3}$ & $0.012$ & $0.080$ & $0.40$ \\
$p$, + feasible seeding & $<10^{-4}$ & $0.026$ & $0.24$ & $1.00$ \\
\bottomrule
\end{tabular}
\end{table}

Unmodified evolutionary search beats blind random search significantly at the three easier levels. At the hardest level it is ahead ($7.8\%$ vs.\ $0.6\%$), but not significantly at $n=8$. Neither seeding variant helps at any level, and each is below the unmodified loop everywhere. An earlier, pre-fix version of this appendix reported that evolution lost at the hardest level ($0.0\%$ vs.\ $1.3\%$), traced that loss to the seeding rule, and found that the two variants closed it at a cost elsewhere. Under the corrected geometry and recalibrated specifications that loss does not occur, so the seeding diagnosis and its trade-off no longer apply. The measurement is geometric, not a rollout. At the two harder levels, 2--6 of 8 seeds per method never reach a feasible design, so best-fitness numbers there are not a fitness comparison.

\subsection{Stricter Validity Criterion and Careful Hand-Coded Editors}
\label{sec:stricter-validity}

The main validity criterion (Section~\ref{sec:results-validity}) only asks whether a compiled body loads and steps without producing non-finite state -- a body that loads and sits motionless passes just as well as one that walks. We test a strictly harder criterion: load-and-step \emph{and} travel at least $2$cm along the forward axis under the universal CPG gait, over the same $1000$-step horizon. We deliberately do not also require the body to stay above the torso-height "fall" floor used elsewhere, since that floor is a soft fitness penalty, not a hard failure condition, and roughly half of genuinely displacing bodies dip below it at some point regardless of mutation strategy.

Table~\ref{tab:stricter-validity} (400 trials per condition, post-fix geometry) shows grammar-guided mutation passing on $370/400$. That is numerically \emph{behind} range-constrained resampling ($377$) and mild-only rescaling ($373$) and level with pure re-initialization ($368$); none of these differences is significant (grammar vs.\ range-constrained: Fisher $p=0.39$). The locomotion bar removes only $6$--$8\%$ of the bodies each of these four conditions produces and does not reorder them. It removes more from the disruptive alternatives (delete element: $120\to75$), and the three most disruptive stay at $0/400$. So the stricter criterion separates the grammar only from alternatives the main criterion already flagged, not from the one it could not separate.

\begin{table}[h]
\caption{Stricter, locomotion-based validity criterion vs.\ loads-and-steps, 400 trials each.}
\label{tab:stricter-validity}
\centering
\small
\begin{tabular}{lcc}
\toprule
Condition & Loads+steps & Loads+steps \emph{and} locomotes $\geq 2$cm \\
\midrule
Range-constrained resampling & 400 & 377 \\
Mild-only rescale & 400 & 373 \\
Grammar-guided (ours) & 400 & 370 \\
Pure re-initialization (sanity floor) & 400 & 368 \\
Adversarial mixture & 326 & 293 \\
Uniform resampling (structure-blind) & 307 & 262 \\
Adversarial mixture, 5 compounded mutations & 276 & 236 \\
Extreme-only (collapse/flip/blow-up) & 246 & 195 \\
Adversarial mixture, 10 compounded mutations & 209 & 177 \\
Delete element & 120 & 75 \\
Subtree-swap crossover & 58 & 53 \\
Naive numeric-average crossover & 46 & 37 \\
Duplicate element & 0 & 0 \\
Byte-level bit-flip & 0 & 0 \\
Latent-space perturbation (GLSO-style contrast) & 0 & 0 \\
\bottomrule
\end{tabular}
\end{table}

Every direct-edit alternative in Table~\ref{tab:validity21} is deliberately naive about structural invariants. A skeptical reader can object that a sufficiently careful, non-adversarial hand-coded editor might close the gap entirely, in which case the grammar's advantage would be convenience rather than necessity. We wrote a genuinely best-effort structural editor: \emph{careful duplication} deep-copies a random subtree, renames every element inside the copy to stay globally unique, and fixes up the matching actuator entry; \emph{careful deletion} removes a random subtree and explicitly removes any actuator entry that referenced a joint inside it.

At $n=400$ (post-fix geometry), careful duplication now also falls short of the grammar ($324/400$), failing on a repeated free-joint name the rename-and-fixup step did not anticipate -- evidence toward necessity, not mere convenience. Careful deletion, despite the same level of care, falls further short ($243/400$), failing on a minimum mass/inertia requirement for a moving body -- a different structural invariant than the one careful duplication was written to avoid, and one that even a careful attempt misses.

\begin{table}[h]
\caption{Counterfactual structural-validity test: careful vs.\ naive hand-coded editors, 400 trials each (post-fix geometry).}
\label{tab:counterfactual}
\centering
\begin{tabular}{lcc}
\toprule
Edit type & Careful (fix-up applied) & Naive (Table~\ref{tab:validity21}) \\
\midrule
Duplication & $324/400$ (falls short of grammar) & $0/400$ \\
Deletion & $243/400$ (falls short of grammar) & $113/400$ \\
\bottomrule
\end{tabular}
\end{table}

The grammar's validity guarantee is pure convenience for some structural edits and genuine necessity for others, within the same experiment -- a more precise claim than either "always necessary" or "a careful implementation always suffices."

\subsection{Validity Gap vs. Task Fitness (Exploratory)}
\label{sec:validity-vs-fitness}

A validity gap only matters downstream if it changes what a search finds. We ran the same $(\mu{+}\lambda)$ loop once mutating grammar genomes and once mutating raw compiled body descriptions with a realistic mix of the Table~\ref{tab:validity21} edit operators, with no feasibility gate ($n=8$ seeds, budget 40). The validity gap reproduces cleanly ($100\%$ vs.\ $58.8\%$ of evaluations valid), but best task fitness does not track it: the raw-edit alternative is directionally ahead, $0.769\pm0.829$ vs.\ $0.351\pm0.109$, with the grammar winning 3 of 8 seeds (paired $t=-1.304$, $p=0.23$). At this sample size the comparison is exploratory and establishes no fitness difference in either direction; a plausible but untested reading is that some loadable raw edits reach unusual, occasionally very mobile body shapes outside the grammar's narrower genome range.

\subsection{Reproducibility: A Simulator-Version Sensitivity}
\label{sec:repro}

Replicating the search comparison on a second machine changed the eight-seed statistic substantially (4/8 seeds won rather than 7, $p=0.576$ rather than $0.082$ for a two-method comparison) despite byte-identical code and seeds. The only difference was the installed physics simulator's minor version; matching the version reproduced the original result to every decimal place across all eight seeds, isolating the simulator's minor-version-dependent solver defaults as the entire cause. We flag this as an open methodological consideration for anyone reproducing these statistics on a different installation, rather than assume a given magnitude transfers unchanged across simulator versions.

\section{Proxy Fitness vs.\ RL-Trained Performance}
\label{app:proxy-rl}

The search fitness, centre-of-mass displacement under a fixed sinusoidal gait, assumes that bodies which move further without learning also do better once a controller is trained. We test this on a catalogue of geometrically feasible bodies. For each body we compute the proxy and an RL-trained fitness, using a tanh MLP policy trained by antithetic evolution strategies on MuJoCo MJX under the identical reward, so that the comparison isolates what training adds.

\begin{figure}[h]
\centering
\includegraphics[width=0.72\linewidth]{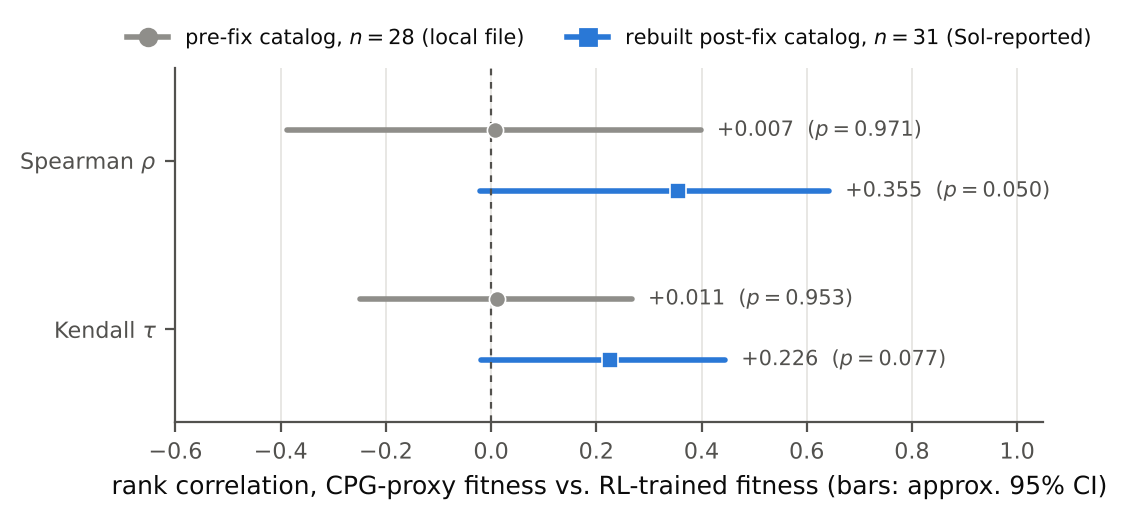}
\caption{Proxy--RL rank correlation on the pre-fix catalogue (local per-body data) and the rebuilt catalogue (cluster-reported coefficients; no per-body data available locally). Bars are approximate 95\% CIs on the Fisher-$z$ scale. Both runs predate the \panini-specific RL-rescoring job (pending, Appendix~\ref{app:rl-rescoring}).}
\label{fig:proxy-rl}
\end{figure}

The first version of this test used a 28-body catalogue built before the limb-mounting geometry fix and found no correlation (Spearman $\rho=0.007$, $p=0.971$; Kendall $\tau=0.011$, $p=0.953$). The catalogue was then rebuilt under the corrected geometry. 28 bodies were drawn from 71 random genomes, which corresponds to a 39\% geometric pass rate against 70\% before the fix, because correctly mounted limbs fail the specification more often. Three \panini final designs were added, for 31 bodies. On this catalogue the correlation is positive but borderline: Spearman $\rho=0.3548$ ($p=0.0501$) and Kendall $\tau=0.2258$ ($p=0.0770$), $n=31$ (Figure~\ref{fig:proxy-rl}). These two values come from the analysis script's terminal output on the compute cluster. The per-body RL results were not transferred back, so we could not re-verify them or draw a scatter plot. We read the result as weak evidence that the proxy carries some information about trained performance. It does not reach conventional significance, and it is too weak to rank bodies on its own. The method-level analysis of \S\ref{sec:rl-rescoring} gives a consistent picture ($\rho=0.17$ across 28 methods).

\section{RL Re-scoring: Supplementary Detail}
\label{app:rl-rescoring}

\begin{figure}[h]
\centering
\includegraphics[width=0.55\linewidth]{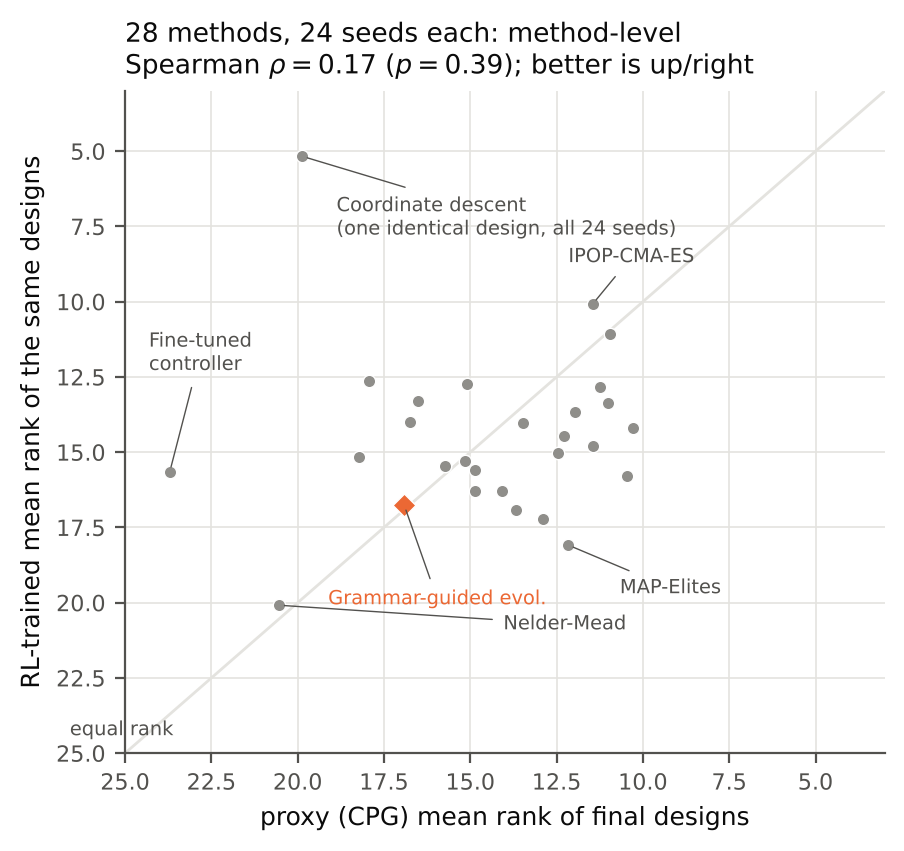}
\caption{Prior run (GGE-based manifest, pre-dates \panini): proxy mean rank against RL mean rank of each method's 24 final designs (28 methods). Coordinate descent's jump comes from a single design repeated across all seeds.}
\label{fig:rl-rescoring}
\end{figure}

\textbf{Prior run (GGE-based manifest).} 672 bodies, all geometrically feasible under the corrected geometry, all loading and stepping. This manifest was built from GGE's final-design run and predates \panini, so \panini is not among the 28 methods below. RL mean ranks, fitness, and the Friedman statistic come from the analysis output. Its proxy-matched-rank field is NaN for every method because of an unresolved bug in the matching step, so we do not use it. The proxy ranks in Table~\ref{tab:rl28} and Figure~\ref{fig:rl-rescoring} are computed instead from each body's re-run proxy fitness in the manifest. The Nemenyi values of \S\ref{sec:rl-rescoring} were first obtained on the cluster; Nemenyi depends only on mean ranks, $k$, and $n$, and recomputing it from the recorded mean ranks reproduces them (for example, $p=0.0004$ for coordinate descent and $p=0.505$ for IPOP-CMA-ES against GGE).

\textbf{\panini-specific re-run.} 144 bodies (\panini, GGE, IPOP-CMA-ES, the simplified CMA-ES, PSO, the cross-entropy method; 24 seeds each), all geometrically feasible, independently re-validated in MuJoCo. All 144 converged under RL training, so, unlike the prior 28-method run, there is no missing-shard or single-degenerate-design distortion. Table~\ref{tab:rl6} gives the full result: Friedman $\chi^2=2.95$, $p=0.71$ over 6 methods and 24 seed-blocks -- no pairwise Nemenyi test is meaningful given the omnibus test alone does not reject the null, so none is reported. \panini ranks 2nd of 6 by RL mean rank, tied in practice with every other method.

\begin{table}[h]
\caption{\panini-specific RL re-scoring: 144 final designs (6 methods $\times$ 24 seeds), sorted by RL mean rank. Friedman $\chi^2=2.95$, $p=0.71$ -- not significant, so no post hoc test is reported.}
\label{tab:rl6}
\centering
\small
\begin{tabular}{lccc}
\toprule
Method & RL mean rank & RL mean fitness & RL std \\
\midrule
PSO & 3.12 & 1.384 & 0.517 \\
\panini\ (proposed) & 3.29 & 1.476 & 0.697 \\
Grammar-guided evolution (GGE) & 3.33 & 1.390 & 0.704 \\
Cross-entropy method & 3.62 & 1.249 & 0.564 \\
IPOP-CMA-ES & 3.75 & 1.179 & 0.449 \\
CMA-ES (simplified) & 3.88 & 1.273 & 0.751 \\
\bottomrule
\end{tabular}
\end{table}

\begin{figure}[h]
\centering
\includegraphics[width=0.6\linewidth]{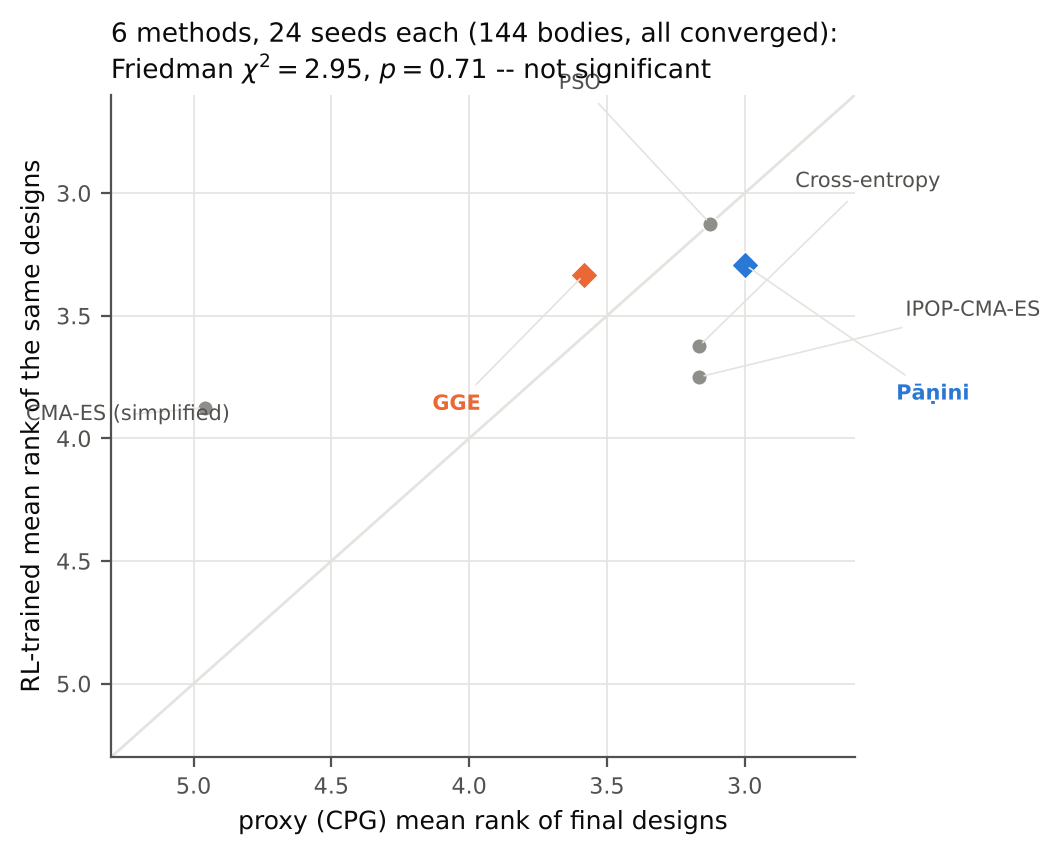}
\caption{\panini-specific re-run: proxy mean rank against RL mean rank of each method's 144 final designs (6 methods, matched-subset proxy ranks). No significant differences (Friedman $p=0.71$); \panini and GGE sit together near the strong end of this small field.}
\label{fig:rl-rescoring-panini}
\end{figure}

\section{Grammar-Native Search Investigations: Compact Summary}
\label{app:edge}

All variants build on \panini's code path and change only the stated mechanism, with the same harness, seeds, and budget accounting as Section~\ref{sec:results-search}; comparisons are Friedman followed by Nemenyi over the listed methods. Hetero runs use a per-limb genome for which only CEM, IPOP-CMA-ES, and PSO have analogues. The curriculum variant is reported in Section~\ref{sec:curriculum}. Other mechanisms listed in Section~\ref{sec:edge} (quality-diversity elite selection, motif reuse, tempering over complexity, constraint-aware repair) did not separate from plain \panini or the reference cluster and are not detailed here.

\begin{table}[h]
\caption{Compact summary of the main non-curriculum investigations.}
\label{tab:edge}
\centering
\scriptsize
\setlength{\tabcolsep}{3pt}
\begin{tabular}{@{}>{\raggedright\arraybackslash}p{3.2cm}>{\raggedright\arraybackslash}p{2.4cm}>{\raggedright\arraybackslash}p{6.0cm}>{\raggedright\arraybackslash}p{2.8cm}@{}}
\toprule
Intervention & Setting & Result & Test \\
\midrule
Larger, heterogeneous design space & hetero, $n$=10, $b$=80 & 3rd of 4: $0.846$ vs.\ PSO $0.927$, IPOP $0.903$ & Friedman $p=0.056$ \\
Fitness noise (30\% std) & $n$=24, $b$=60 & $0.301$ vs.\ CEM $0.305$; $n$=6 screening signal not reproduced & Friedman $p=0.39$ \\
Subtree crossover/mutation & hetero, $n$=24, $b$=60/120 & 2nd of 5 at both budgets ($0.841$, $0.945$); IPOP $0.829$, $0.974$; structurally new offspring 94\% of the time & Friedman $0.39$/$0.034$; Nemenyi $\ge0.055$ \\
Transfer to held-out limb counts & hetero, $n$=24, $b$=80 & Last of 5: $0.810$ vs.\ IPOP $0.960$, PSO $0.951$, cold \panini $0.893$ (17/24 seeds worse than cold control) & Nemenyi $0.012$, $0.038$ (worse) \\
Niche regimes ($b$=10--35; stricter specs) & 8 screens, $n$=8/24 & Best screen ($b$=35, $0.375$ vs.\ CEM $0.355$) does not survive correction & Friedman $p=0.51$; Holm 0/8 \\
\bottomrule
\end{tabular}
\end{table}

\section{Closed Kinematic Chains}
\label{app:closed-chain}

Extending the morphology grammar to branching limbs needs only a new rule following the existing required-role gating pattern; extending it to closed kinematic chains needs a genuinely new verification step, which this appendix characterizes directly by building the simplest non-trivial closed chain: a planar four-bar linkage (a ground link of length $L_0$ connecting two torso mount points, a crank of length $L_1$ driven by angle $\theta_1$, a rocker of length $L_3$ at angle $\theta_3$, and a coupler of length $L_2$ that must physically connect the crank's tip to the rocker's tip).

The crank and rocker are each ordinary open-chain limb derivations from the torso -- exactly the branching-limb case the grammar already handles. What is new: in a closed chain, $\theta_3$ is not a free genome parameter. It is \emph{determined} by the other three link lengths and $\theta_1$ via the loop-closure constraint $|\mathrm{crank\_tip}(\theta_1) - \mathrm{rocker\_tip}(\theta_3)| = L_2$, which the inheritance mechanism -- designed to carry a value forward from an earlier rule, not to solve a constraint between two independently-derived subtrees' endpoints -- has no analogue for.

We solve this constraint numerically (bisection over $\theta_3 \in [0, 2\pi)$, refined with Brent's method, $x_\mathrm{tol}=10^{-12}$) for 500 randomly sampled configurations, link lengths drawn from the same range as this paper's own segment-length bound ($[0.05, 0.30]$).

\textbf{Assembly feasibility is a real, non-trivial question.} Only $318/500$ ($63.6\%$) of randomly sampled link-length combinations admit any real solution to the loop-closure equation at all -- the remaining $36.4\%$ are not closeable by \emph{any} $\theta_3$, a classical consequence of four-bar linkage geometry (related to the Grashof condition) that the grammar's existing parameter clamping has no mechanism to detect or avoid.

\textbf{Where a solution exists, the verification step itself is sound.} For every closeable configuration, we check the claim that two derivation paths must agree on a shared joint's position, by building the coupler at the \emph{solved} length $L_2$ and measuring how far its endpoint lands from the crank's own tip. The measured residual is mean $1.5\times10^{-15}$, max $6.3\times10^{-14}$ -- floating-point precision, to the same standard of a measured, bounded defect this paper applies to $\Phi$'s own homomorphism defect (\S\ref{sec:results-homomorphism}).

A closed chain therefore requires (a) an assembly-feasibility check with no analogue in the current grammar, and (b) a numerical verification step categorically different from inheritance or specificity arbitration. We integrate both into the grammar directly, rather than leaving the solver standalone, through one additive extension: a new rule whose parameters are computed by solving a constraint over already-derived state, after the normal inheritance/override merge, instead of only from defaults or inheritance. Every existing rule is unaffected by this extension and the full test suite still passes unchanged. The rocker's mount angle becomes this rule's resolved output rather than a free parameter, and an unsolvable configuration is rejected through the grammar's own existing failure path for an inapplicable rule, with no new exception type introduced.

We benchmark this integration on 300 random four-bar configurations, comparing the grammar-wrapped rule (with a bounded repair loop that shrinks the coupler/rocker lengths and re-derives, up to 5 attempts, on an initial failure) against the standalone solver used as-drawn with no repair. The two paths agree on the as-drawn configuration $92.7\%$ of the time, confirming that wrapping the solver as a grammar rule changes nothing about its correctness; the grammar-wrapped path's repair loop raises the overall closeable rate from $63.0\%$ to $70.3\%$ ($7.3\%$ repaired, $29.7\%$ rejected as unfixable within 5 attempts). The forward-kinematic residual at the solved angle remains at floating-point precision (mean $1.6\times10^{-15}$, max $6.3\times10^{-14}$), matching the homomorphism-defect standard of \S\ref{sec:results-homomorphism}. Beyond what the standalone solver could check at all, we also verify genuine downstream simulation: compiling the solved linkage with a native rigid-body equality constraint joining the crank and coupler tips, every closeable configuration loads and steps 100 physics steps without blowing up, with the equality constraint's residual under real dynamics averaging $3.65\times10^{-3}$ (max $3.98\times10^{-2}$) -- small, but non-zero in a way the offline forward-kinematics check, which never touches a physics engine, cannot see. Wiring the solver into the rule-application machinery is therefore no longer future work; what remains open is generalizing this mechanism beyond this one hand-written constraint to a general library of closed-chain topologies.

\section{Manipulation Benchmark: Full Results}
\label{app:manip-benchmark}

Section~\ref{sec:manip-benchmark} summarizes this result; this appendix gives the full setup and table. The benchmark is built around the existing Approach$\to$Grasp$\to$Lift$\to$Transport$\to$Place rule chain. The end effector is a kinematically-driven point (with no actuators or joint limits of its own, so it isolates whether the \emph{staged plan} succeeds from whether a real controller can track it); the object is a free dynamic body under gravity; grasping is a rigid kinematic pin between object and end-effector pose, engaged within a tolerance radius of the object and released at the Place target -- a simplified-grasp proxy with no finger-contact dynamics, chosen for stability under the end effector's discontinuous-looking position updates.

Three tasks of increasing length are run on the same 150 random scenes for the grammar-derived plan and two baselines of deliberately different strength. ``Baseline'' moves in one straight line from the end-effector's start pose directly to the final target, with no knowledge of the object's location, attempting a grasp only opportunistically if that line happens to pass near the object. ``Informed'' is handed the exact same ground-truth object and target positions the grammar's own waypoints are computed from and deliberately routes through the object before the target (start-to-object, then object-to-target, same total step budget), but still lacks the grammar's staged ordering -- no separate hover-above waypoint, no separate grasp-depth waypoint, no separate lift-to-height waypoint before transporting. Both baselines attempt a grasp via the same tolerance-triggered mechanism as the grammar, so any gap is attributable to staging, not to a different grasp mechanism or time budget.

\begin{table}[h]
\caption{Manipulation benchmark, $n=150$ random scenes per cell. ``Reach'' success is reaching within hover range of the object (grammar and baseline target the same hover point, so that cell is a sanity check, not a differentiator); the other two tasks require an actual grasp.}
\label{tab:manip-benchmark}
\centering
\small
\begin{tabular}{llccc}
\toprule
Task (fixed horizon) & Planner & Success & Grasp rate & Mean place error \\
\midrule
Reach (120 steps) & Grammar & 86.0\% & --- & --- \\
Reach (120 steps) & Baseline (uninformed) & 86.0\% & --- & --- \\
Reach (120 steps) & Informed & \textbf{100.0\%} & --- & --- \\
Grasp+Lift (360 steps) & Grammar & \textbf{100.0\%} & 100.0\% & --- \\
Grasp+Lift (360 steps) & Baseline (uninformed) & 0.0\% & 0.0\% & --- \\
Grasp+Lift (360 steps) & Informed & \textbf{100.0\%} & 100.0\% & --- \\
Place (600 steps) & Grammar & \textbf{100.0\%} & 100.0\% & 0.00\,cm \\
Place (600 steps) & Baseline (uninformed) & 0.0\% & 0.0\% & 52.91\,cm \\
Place (600 steps) & Informed & \textbf{100.0\%} & 100.0\% & 1.71\,cm \\
\bottomrule
\end{tabular}
\end{table}

No simulation failures occur in any of the $150\times3\times3=1350$ trials. Against the uninformed baseline the gap is total: it never grasps the object across 150 scenes, because a straight line from start to final goal essentially never happens to pass within grasp tolerance of an object that is not on that line, so it arrives carrying nothing -- a direct physical consequence of having no explicit grasp waypoint, but one that does not yet separate ``knows nothing about the task'' from ``lacks staged ordering specifically.'' Against the informed baseline, which is told exactly where the object and target are and deliberately visits the object, that separation is clean, and the result reverses what the uninformed comparison alone would suggest: Grasp+Lift and Place tie at $100\%$ success for both, and on Reach the informed planner is slightly \emph{ahead} ($100\%$ vs.\ $86\%$), because it is routed directly at the object itself rather than at a hover point above it. The grammar's $0.00$\,cm mean place error is close to tautological -- its place target \emph{is} the support location the error is measured against -- while the informed planner's non-zero but still small $1.71$\,cm reflects a real (if modest) placement error under the same kinematic grasp mechanism. The honest conclusion is narrower than the uninformed-baseline comparison alone would support: on this benchmark, the grammar's explicit staging is not shown to outperform a competent, object-aware but unstaged planner once that planner is told where to look; it is a proof-of-concept that the staged derivation is at least as capable as a reasonable alternative, not a demonstrated advantage over a strong competitor.

\section{Supplementary Figures}
\label{app:figs}

\begin{figure}[h]
\centering
\includegraphics[width=0.55\linewidth]{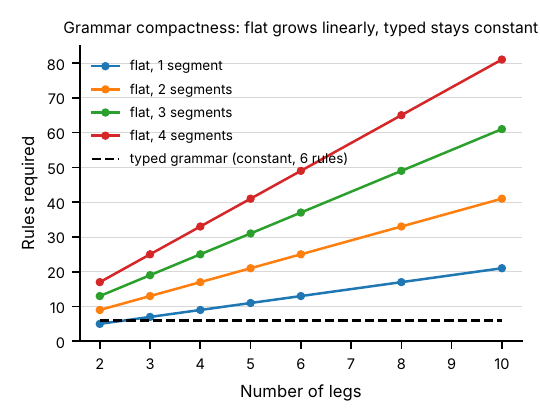}
\caption{Flat rule count grows linearly with body size at every segment count tested; typed count is constant.}
\label{fig:compactness}
\end{figure}

\begin{figure}[h]
\centering
\includegraphics[width=0.5\linewidth]{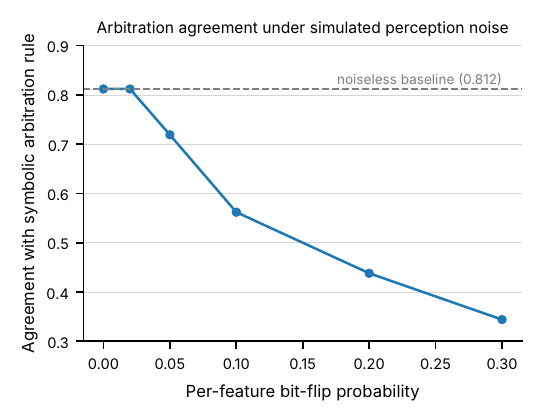}
\caption{Containment-based arbitration degrades smoothly, not catastrophically, under simulated perception noise.}
\label{fig:noise}
\end{figure}

\section{Validity Protocol in the Very-Low-Feasibility Environment: Supplementary Results}
\label{app:validity-hard}

Parents are feasible, perfectly symmetric 8-limb grammar bodies; every condition edits the same parent. Joint feasibility is computed on the compiled, forward-settled body; its verdict matches the tree-based check on all 1300 test bodies at every level.

\begin{table}[t]
\caption{The 15 validity conditions of Table~\ref{tab:validity21} in the very-low-feasibility environment ($n_\ell=8$ limbs, 300 trials per cell). Every condition edits the same feasible parent and is scored after 3 edits, the original protocol. ``Legal'': loads and steps. Feasible: also passes the joint spec; columns are pooled target pass rates. $^{\dagger}$Edits the body in only $15\%$ of trials (Table~\ref{tab:new-feasible}). $^{\ddagger}$Interpolates between two feasible, grammar-built parents.}
\label{tab:validity-hard}
\centering
\small
\begin{tabular}{lccccc}
\toprule
& Legal & \multicolumn{4}{c}{Feasible} \\
Condition & & $8\%$ & $3\%$ & $1\%$ & $0.3\%$ \\
\midrule
Naive numeric-average crossover$^{\ddagger}$ & 300 & 300 & 300 & 300 & 300 \\
Mild-only rescale$^{\dagger}$ & 300 & 296 & 294 & 294 & 282 \\
\textbf{Grammar-guided (ours)} & \textbf{300} & \textbf{285} & \textbf{266} & \textbf{225} & \textbf{169} \\
Adversarial mixture & 250 & 212 & 199 & 193 & 184 \\
Adversarial mixture, 5 compounded & 234 & 194 & 186 & 177 & 162 \\
Range-constrained resampling & 300 & 166 & 154 & 150 & 129 \\
Extreme-only (collapse/flip/blow-up) & 200 & 140 & 133 & 125 & 117 \\
Uniform resampling (structure-blind) & 230 & 108 & 93 & 90 & 67 \\
Adversarial mixture, 10 compounded & 165 & 97 & 83 & 74 & 59 \\
Pure re-initialization (sanity floor) & 300 & 47 & 20 & 8 & 3 \\
Delete element & 29 & 28 & 20 & 14 & 3 \\
Subtree-swap crossover & 12 & 10 & 9 & 8 & 8 \\
Duplicate element; bit-flip; latent (GLSO-style), each & 0 & 0 & 0 & 0 & 0 \\
\bottomrule
\end{tabular}
\end{table}

Naive numeric-average crossover passes $300/300$ because it interpolates between two feasible parents, here both built by the grammar; it cannot reach a design outside the segment between them, and from scratch it has no feasible parents to average. Table~\ref{tab:new-feasible} below counts only trials that produce a genuinely new feasible body, the fairer comparison Section~\ref{sec:validity-hard} cites.

\begin{table}[h]
\caption{Flat per-limb parameterization vs.\ grammar construction, $n_\ell=8$ limbs. Pass rates from 6000 fresh samples (flat 95\% Wilson upper bound $0.06\%$ in every row); search: 200 evaluations, 20 seeds, CMA-ES with a constraint-violation objective.}
\label{tab:hard-space}
\centering
\small
\begin{tabular}{lcccc}
\toprule
Pooled target pass rate & $8\%$ & $3\%$ & $1\%$ & $0.3\%$ \\
\midrule
Sampling, flat & $0.0\%$ & $0.0\%$ & $0.0\%$ & $0.0\%$ \\
Sampling, grammar & $13.1\%$ & $6.5\%$ & $2.8\%$ & $1.0\%$ \\
CMA-ES, flat: seeds with any feasible design & $0/20$ & $0/20$ & $0/20$ & $0/20$ \\
CMA-ES, grammar: seeds with any feasible design & $20/20$ & $20/20$ & $20/20$ & $20/20$ \\
CMA-ES, grammar: feasible evaluations & $30.7\%$ & $31.1\%$ & $27.4\%$ & $26.9\%$ \\
\bottomrule
\end{tabular}
\end{table}

\begin{table}[h]
\caption{Share of 120 trials, after 3 edits, that produce a \emph{new} (leg tips moved $>1$\,mm) and jointly feasible body, with mean leg-tip displacement from the parent; same environment as Table~\ref{tab:validity-hard}.}
\label{tab:new-feasible}
\centering
\small
\begin{tabular}{lccccc}
\toprule
& \multicolumn{4}{c}{New and feasible} & Tip displacement \\
Condition & $8\%$ & $3\%$ & $1\%$ & $0.3\%$ & ($8\%$ level) \\
\midrule
\textbf{Grammar-guided (ours)} & \textbf{$77\%$} & \textbf{$65\%$} & \textbf{$52\%$} & \textbf{$37\%$} & $15.3$\,mm \\
Range-constrained resampling & $18\%$ & $13\%$ & $12\%$ & $11\%$ & $90.0$\,mm \\
Mild-only rescale$^{\dagger}$ & $14\%$ & $14\%$ & $12\%$ & $9\%$ & $0.4$\,mm \\
Adversarial mixture & $12\%$ & $8\%$ & $7\%$ & $4\%$ & $32.0$\,mm \\
Naive numeric-average crossover$^{\ddagger}$ & $97\%$ & $96\%$ & $97\%$ & $83\%$ & $51.5$\,mm \\
\bottomrule
\end{tabular}
\end{table}

\begin{table}[h]
\caption{Topology change: the 8-limb spec is held fixed and every method starts from the same 6-limb body (lowest constraint violation among 2000 grammar 6-limb bodies). 200 evaluations, 20 seeds, $(\mu{+}\lambda)$ search minimising constraint violation.}
\label{tab:topology}
\centering
\small
\begin{tabular}{lcccc}
\toprule
Method & $8\%$ & $3\%$ & $1\%$ & $0.3\%$ \\
\midrule
Range-constrained resampling: seeds feasible & $0/20$ & $0/20$ & $0/20$ & $0/20$ \\
Per-field clip (flat): seeds feasible & $0/20$ & $0/20$ & $0/20$ & $0/20$ \\
Grammar (5-parameter genome): seeds feasible & $20/20$ & $20/20$ & $20/20$ & $20/20$ \\
Grammar: feasible evaluations & $69.6\%$ & $64.5\%$ & $63.0\%$ & $62.4\%$ \\
\bottomrule
\end{tabular}
\end{table}

\section{Objection Experiments: Full Results}
\label{app:objections}

\begin{table}[h]
\caption{Four objections in the hard space ($n_\ell=8$), easiest and hardest pooled target pass rate. Each cell is baseline vs.\ grammar; (1) uses per-field clipping, (2) CMA-ES from scratch, (4) the grammar generator.}
\label{tab:objections}
\centering
\small
\begin{tabular}{llcc}
\toprule
Objection & Measure (baseline vs.\ grammar) & $8\%$ & $0.3\%$ \\
\midrule
(1) & Feasible after 10 mutations & $30.0$ vs.\ $86.7\%$ & $10.7$ vs.\ $27.3\%$ \\
(1) & Best material in search (lower better) & $0.678$ vs.\ $0.412$ & $0.821$ vs.\ $0.630$ \\
(2) & Box feasible (flat 64-D vs.\ 5-D) & $0.4$ vs.\ $19.3\%$ & $0.1$ vs.\ $35.5\%$ \\
(2) & Seeds feasible (flat vs.\ 12-D grammar) & $0/20$ vs.\ $20/20$ & $0/20$ vs.\ $6/20$ \\
(3) & Add a limb (hand-coded vs.\ grammar) & $0$ vs.\ $100\%$ & $0$ vs.\ $100\%$ \\
(4) & Physics calls per feasible (gate off vs.\ on) & $47.6$ vs.\ $1$ & $857$ vs.\ $1$ \\
\bottomrule
\end{tabular}
\end{table}

\begin{table}[h]
\caption{Coupled, cross-limb feasibility: support-polygon area threshold calibrated on 300 fresh genomes to target pass rate $p$; jointly feasible mutants out of 400 per cell (post-fix geometry). Brackets: $\chi^2$ $p$ against grammar-guided where $p<0.05$. All 400 mutants in every row loaded and stepped.}
\label{tab:coupled-feasibility}
\centering
\small
\begin{tabular}{lccc}
\toprule
Condition & $p\approx0.30$ (area$\geq0.099$) & $p\approx0.10$ ($\geq0.127$) & $p\approx0.05$ ($\geq0.142$) \\
\midrule
Grammar-guided (ours) & 94 & 48 & 24 \\
Pure re-initialization & 78 & 37 & 35 \\
Independent random mount angle & 49 [$<10^{-4}$] & 21 [$0.001$] & 10 [$0.023$] \\
Range-constrained resampling & 103 & 52 & 41 [$0.038$] \\
Hand-coded, coordinated (no grammar) & 87 & 38 & 32 \\
Hand-coded, random angle (no grammar) & 42 [$<10^{-4}$] & 18 [$<10^{-3}$] & 13 \\
\bottomrule
\end{tabular}
\end{table}

All experiments use the hard design space of Section~\ref{sec:hard-design-space} at $n_\ell=8$ limbs and the four pooled-calibrated difficulty levels. Feasible parents are perfectly symmetric grammar bodies, and the same parent is handed to every method. The flat mutation operators perturb or redraw one of the $64$ normalised fields and clip it to its own range; grammar mutation perturbs one grammar parameter and re-derives the body. The search objective is total leg material $\sum \pi r^2 \ell$ over all built segments; body mass is not used because in this simulator it depends only on segment count.

\begin{table}[h]
\caption{Objection 1, random walk: fraction of 150 walks still feasible after $k$ unselected mutations from the same feasible parent, $n_\ell=8$ limbs. Columns are pooled target pass rates.}
\label{tab:obj-walk}
\centering
\small
\begin{tabular}{llcccc}
\toprule
Mutation & $k$ & $8\%$ & $3\%$ & $1\%$ & $0.3\%$ \\
\midrule
Per-field clip & 1 & $89.3\%$ & $85.3\%$ & $92.7\%$ & $77.3\%$ \\
 & 5 & $50.0\%$ & $51.3\%$ & $55.3\%$ & $32.7\%$ \\
 & 10 & $30.0\%$ & $26.0\%$ & $25.3\%$ & $10.7\%$ \\
 & 20 & $10.7\%$ & $5.3\%$ & $4.7\%$ & $4.0\%$ \\
 & 50 & $0.0\%$ & $0.0\%$ & $0.7\%$ & $0.0\%$ \\
\midrule
Per-field resample & 1 & $81.3\%$ & $67.3\%$ & $48.7\%$ & $38.7\%$ \\
 & 5 & $30.0\%$ & $8.7\%$ & $4.0\%$ & $2.0\%$ \\
 & 10 & $8.7\%$ & $1.3\%$ & $0.0\%$ & $0.0\%$ \\
 & 20 & $0.7\%$ & $0.0\%$ & $0.0\%$ & $0.0\%$ \\
 & 50 & $0.0\%$ & $0.0\%$ & $0.0\%$ & $0.0\%$ \\
\midrule
Grammar & 1 & $96.7\%$ & $96.0\%$ & $91.3\%$ & $81.3\%$ \\
 & 5 & $90.0\%$ & $82.7\%$ & $61.3\%$ & $47.3\%$ \\
 & 10 & $86.7\%$ & $74.7\%$ & $42.0\%$ & $27.3\%$ \\
 & 20 & $76.0\%$ & $52.0\%$ & $23.3\%$ & $6.0\%$ \\
 & 50 & $46.0\%$ & $20.7\%$ & $5.3\%$ & $0.0\%$ \\
\bottomrule
\end{tabular}
\end{table}

\begin{table}[h]
\caption{Objection 1, budgeted search from the same feasible parent (200 evaluations, 20 seeds, $n_\ell=8$): minimise leg material subject to joint feasibility. ``Best'' is the best feasible material relative to the parent (lower is better); ``Feas.'' is the share of evaluations that were feasible.}
\label{tab:obj-search}
\centering
\small
\begin{tabular}{lcccccccc}
\toprule
& \multicolumn{2}{c}{$8\%$} & \multicolumn{2}{c}{$3\%$} & \multicolumn{2}{c}{$1\%$} & \multicolumn{2}{c}{$0.3\%$} \\
Method & Best & Feas. & Best & Feas. & Best & Feas. & Best & Feas. \\
\midrule
Flat, per-field clip ES ($\sigma=0.1$) & $0.876$ & $87.2\%$ & $0.883$ & $84.8\%$ & $0.887$ & $82.8\%$ & $0.903$ & $77.6\%$ \\
Flat, per-field clip ES ($\sigma=0.3$) & $0.731$ & $81.5\%$ & $0.765$ & $76.8\%$ & $0.781$ & $74.9\%$ & $0.821$ & $70.4\%$ \\
Flat, per-field resample ES & $0.678$ & $67.2\%$ & $0.730$ & $57.0\%$ & $0.772$ & $49.2\%$ & $0.828$ & $43.8\%$ \\
Flat, CMA-ES ($\sigma=0.1$) & $1.000$ & $0.5\%$ & $1.000$ & $0.5\%$ & $1.000$ & $0.5\%$ & $1.000$ & $0.5\%$ \\
Flat, CMA-ES ($\sigma=0.01$) & $0.992$ & $7.8\%$ & $0.992$ & $7.8\%$ & $0.994$ & $7.1\%$ & $0.995$ & $4.5\%$ \\
\midrule
Grammar ES & $0.523$ & $94.9\%$ & $0.543$ & $89.2\%$ & $0.568$ & $86.8\%$ & $0.630$ & $82.5\%$ \\
Grammar CMA-ES & $0.412$ & $69.7\%$ & $0.410$ & $64.3\%$ & $0.525$ & $53.4\%$ & $0.697$ & $34.6\%$ \\
\bottomrule
\end{tabular}
\end{table}

\begin{table}[h]
\caption{Objection 2: separability and genome size, $n_\ell=8$. Box rows sample uniformly inside the tightest per-field interval box around 400 feasible designs; the rotation-augmented box first rotates each feasible body by a random angle, which never changes feasibility. The last rows run CMA-ES from scratch over a 12-parameter grammar genome (mirror-pair lengths and radii), 200 evaluations, 20 seeds.}
\label{tab:obj-separable}
\centering
\small
\begin{tabular}{lcccc}
\toprule
Test & $8\%$ & $3\%$ & $1\%$ & $0.3\%$ \\
\midrule
Flat 64-D, uniform & $0.0\%$ & $0.0\%$ & $0.0\%$ & $0.0\%$ \\
Flat 64-D, box & $0.4\%$ & $0.5\%$ & $0.2\%$ & $0.1\%$ \\
Flat 64-D, rotation-augmented box & $0.0\%$ & $0.0\%$ & $0.0\%$ & $0.0\%$ \\
5-D genome, uniform & $1.2\%$ & $0.3\%$ & $0.1\%$ & $0.0\%$ \\
5-D genome, box & $19.3\%$ & $19.7\%$ & $26.0\%$ & $35.5\%$ \\
\midrule
12-D grammar, uniform & $0.4\%$ & $0.0\%$ & $0.0\%$ & $0.0\%$ \\
12-D grammar CMA-ES: seeds with a feasible design & $20/20$ & $18/20$ & $15/20$ & $6/20$ \\
12-D grammar CMA-ES: feasible evaluations & $25.7\%$ & $10.2\%$ & $2.5\%$ & $0.2\%$ \\
\bottomrule
\end{tabular}
\end{table}

\begin{table}[h]
\caption{Objection 3: hand-coded editors vs.\ grammar edits on a feasible parent ($n_\ell=8$, 400 trials per cell); feasibility of the edited body.}
\label{tab:obj-editors}
\centering
\small
\begin{tabular}{llcccc}
\toprule
Edit & Editor & $8\%$ & $3\%$ & $1\%$ & $0.3\%$ \\
\midrule
Add limb & Hand-coded & $0.0\%$ & $0.0\%$ & $0.0\%$ & $0.0\%$ \\
 & Hand-coded + re-space & $100.0\%$ & $100.0\%$ & $100.0\%$ & $100.0\%$ \\
 & Grammar & $100.0\%$ & $100.0\%$ & $100.0\%$ & $100.0\%$ \\
\midrule
Remove limb & Hand-coded & $0.0\%$ & $0.0\%$ & $0.0\%$ & $0.0\%$ \\
 & Hand-coded + re-space & $93.0\%$ & $82.5\%$ & $58.8\%$ & $0.0\%$ \\
 & Grammar & $93.0\%$ & $82.5\%$ & $58.8\%$ & $0.0\%$ \\
\midrule
Add segment & Hand-coded & $100.0\%$ & $100.0\%$ & $100.0\%$ & $100.0\%$ \\
 & Grammar & $100.0\%$ & $100.0\%$ & $100.0\%$ & $100.0\%$ \\
\midrule
Lengthen & Hand-coded & $99.2\%$ & $97.5\%$ & $97.5\%$ & $98.8\%$ \\
 & Grammar & $100.0\%$ & $100.0\%$ & $100.0\%$ & $100.0\%$ \\
\bottomrule
\end{tabular}
\end{table}

\begin{table}[h]
\caption{Objection 4: generator $\times$ gate factorial, 6000 draws per cell, $n_\ell=8$. Physics calls per feasible design found (``--'' when none was found).}
\label{tab:obj-factorial}
\centering
\small
\begin{tabular}{lcccc}
\toprule
Generator, gate & $8\%$ & $3\%$ & $1\%$ & $0.3\%$ \\
\midrule
Flat, gate off & -- (0) & -- (0) & -- (0) & -- (0) \\
Flat, gate on & -- (0) & -- (0) & -- (0) & -- (0) \\
Grammar, gate off & $47.6$ (126) & $105.3$ (57) & $315.8$ (19) & $857.1$ (7) \\
Grammar, gate on & $1.0$ (126) & $1.0$ (57) & $1.0$ (19) & $1.0$ (7) \\
\bottomrule
\end{tabular}
\\[2pt]
\footnotesize Parentheses: feasible designs found in 6000 draws.
\end{table}

\section{Full Validity Comparison}
\label{app:validity-table}
The full validity comparison is given in table \ref{tab:validity21}.

\begin{table}[h]
\caption{Validity comparison, grammar-guided mutation vs.\ 14 alternatives, 400 trials each (post-fix geometry)}
\label{tab:validity21}
\centering
\small
\begin{tabular}{lc}
\toprule
Condition & Valid / 400 \\
\midrule
Grammar-guided (ours) & 400 (100\%) \\
Pure re-initialization (sanity floor, not a real competitor) & 400 (100\%) \\
Mild-only rescale & 400 (100\%) \\
Range-constrained resampling & 400 (100\%) \\
Adversarial mixture & 326 (82\%) \\
Uniform resampling (structure-blind) & 307 (77\%) \\
Adversarial mixture, 5 compounded mutations & 276 (69\%) \\
Extreme-only (collapse/flip/blow-up) & 246 (62\%) \\
Adversarial mixture, 10 compounded mutations & 209 (52\%) \\
Delete element & 120 (30\%) \\
Subtree-swap crossover & 58 (14\%) \\
Naive numeric-average crossover & 46 (12\%) \\
Duplicate element & 0 \\
Byte-level bit-flip & 0 \\
Latent-space perturbation (GLSO-style contrast) & 0 \\
\bottomrule
\end{tabular}
\end{table}

\section{Raw Per-Seed Results}
\label{app:rawdata}

Per-seed values for every search comparison in this paper are released with the code; Figure~\ref{fig:boxplot} plots them for the main comparison. The six-method per-seed table of earlier versions predates the geometry fix and is withdrawn.

\begin{table}[h]
\caption{Per-seed best fitness for Table~\ref{tab:gate} (budget 40, post-fix geometry): no gate (None), geometric gate (Geo.), geometric gate plus gait pre-check (+Gait). The earlier 8-seed version of this table predates the geometry fix and is withdrawn.}
\centering
\small
\begin{tabular}{cccc|cccc}
\toprule
Seed & None & Geo. & +Gait & Seed & None & Geo. & +Gait \\
\midrule
0 & $0.468$ & $0.320$ & $-0.082$ & 12 & $0.505$ & $0.264$ & $0.288$ \\
1 & $0.129$ & $0.257$ & $0.346$ & 13 & $0.243$ & $0.385$ & $0.296$ \\
2 & $0.457$ & $0.484$ & $0.523$ & 14 & $0.444$ & $0.557$ & $0.444$ \\
3 & $0.473$ & $0.557$ & $0.360$ & 15 & $0.396$ & $0.291$ & $0.255$ \\
4 & $0.308$ & $-0.590$ & $-0.590$ & 16 & $0.398$ & $0.524$ & $0.380$ \\
5 & $0.283$ & $0.279$ & $0.213$ & 17 & $0.205$ & $0.126$ & $0.110$ \\
6 & $0.337$ & $0.450$ & $0.105$ & 18 & $0.291$ & $0.300$ & $0.451$ \\
7 & $0.354$ & $0.316$ & $0.291$ & 19 & $0.223$ & $0.174$ & $-0.106$ \\
8 & $0.501$ & $0.520$ & $0.431$ & 20 & $0.163$ & $0.323$ & $0.333$ \\
9 & $0.099$ & $0.410$ & $0.436$ & 21 & $0.304$ & $0.164$ & $0.498$ \\
10 & $0.344$ & $0.524$ & $0.429$ & 22 & $0.434$ & $0.520$ & $0.497$ \\
11 & $0.405$ & $0.422$ & $0.328$ & 23 & $0.375$ & $0.434$ & $0.232$ \\
\bottomrule
\end{tabular}
\end{table}

\end{document}